\documentclass[10pt]{article}

\usepackage[preprint]{tmlr}

\usepackage{microtype}
\usepackage{booktabs}
\usepackage{multirow}
\usepackage{placeins}
\usepackage{float}
\usepackage{tabularx}
\usepackage{longtable}
\usepackage{graphicx}
\usepackage{subcaption}
\usepackage{tikz}
\usetikzlibrary{positioning}
\usepackage{amsmath,amssymb}
\usepackage{xcolor}
\usepackage[hidelinks]{hyperref}
\usepackage{url}
\usepackage{enumitem}

\graphicspath{{figures/}}
\newcommand{\dataset}[1]{\textsc{#1}}
\newcommand{\est}[1]{\textsf{#1}}
\newcommand{\V}{V(\pi_e)}
\newcommand{\Vhat}{\widehat{V}}
\newcommand{\DMDRREFIT}{on the exact-value datasets, $5$ seeds $\times$ $80$ cells with
$\hat\mu$ refit inside every bootstrap resample (\texttt{make repro-refit-intervals}):
\est{DM}$-$\est{DR}$=-0.0005$, 95\% CI $[-0.0038,+0.0028]$}
\newcommand{\ALIGNRESULT}{%
Misaligned logging modestly raises every estimator's error but preserves the ordering.
Pooled over the exact-value and RCT datasets ($5{,}040$ estimator rows $=4$ datasets $\times5$ seeds $\times3$ budgets $\times3$ overlaps $\times4$ (logging, target) pairs $\times7$ estimators, the seventh being perturbation-DR, which is computed but excluded from the ordering below)
on \dataset{synthetic}, \dataset{IHDP}, \dataset{Hillstrom} and \dataset{Lenta} (the four
datasets on which both a \est{T-} and an \est{S-learner} logger can be built), median
cell-level relative RMSE rises from $0.0214$ to $0.0240$ for the model-based estimators
(\est{DM}, \est{DR}, \est{Switch-DR}) and from $0.0361$ to $0.0417$ for the IPS family
(\est{IPS}, \est{mIPS}) --- $+12.3\%$ and $+15.4\%$ respectively. The ordering is
preserved at the family level (\est{DR} $0.0239$, \est{DM} $0.0240$, \est{Switch-DR} $0.0243$,
\est{SNIPS} $0.0307$, \est{IPS} $0.0408$, \est{mIPS} $0.0424$; \est{DR} and \est{DM} swap
by $0.0001$, well inside noise, consistent with our refusal to order them), and the model-based
advantage is if anything slightly widened, though the difference in degradation rates is
small enough that we do not claim weighting estimators are materially more
misalignment-sensitive. The RQ1--RQ2 conclusions therefore survive \emph{this} perturbation. We are deliberate
about its strength: logging under one meta-learner and evaluating another perturbs the
logger between two correlated scoring rules, which is weaker than the
candidate-\emph{independent} logger that produces the RQ4 attenuation of
Section~\ref{sec:rq4}. We did not re-score accuracy under that logger, so the RQ1--RQ2
\emph{magnitudes} should be read as self-aligned-logger quantities; what this check
supports is that the ordering is not an artifact of alignment, not that the gap size is
logger-invariant.}

\def\month{MM}
\def\year{2026}
\def\openreview{\url{https://openreview.net/forum?id=XXXX}}

\title{%
  When Can You Trust Offline Evaluation of Equal-Cost Top-$k$ Allocation?\\
  A Controlled, Reproducible Benchmark and Practitioner's Guide
}

\IfFileExists{author.tex}{% Author identity for the NAMED builds (arXiv preprint, camera-ready).
\author{%
  \name Binshuang Li\\
  \addr Independent Researcher
}
}{%
  \author{\name Anonymous Author(s) \email anon@example.com\\
          \addr Anonymous Institution}%
}

\begin{document}

\maketitle
{\let\thefootnote\relax\footnotetext{Large language models assisted with writing and
code; the authors take full responsibility for all content (see the Use of Generative AI
statement).}}

% ============================================================
\begin{abstract}
Organizations routinely decide \emph{whom} to treat under a budget, and want to know what
a targeting rule \emph{would} have earned before deploying it. Off-policy evaluation
promises this from logged data. But the deployable rule is a \emph{deterministic} top-$k$
policy: it removes all averaging over actions, so weak overlap hits the estimate directly.
We benchmark six estimators across five datasets and two known-effect sweeps, and
validate the mechanisms against a non-simulated paired reference.

First, weak overlap is governed by logger--target \emph{action} alignment, not by logging
sharpness alone.
Temperature is not a valid overlap parameter: what governs support is the logger's
probability of the \emph{target's actions}. Over the tested range, sharpening a logger built
from the target's own score barely moves overlap; disagreement at the action level
collapses it. Effective
sample size, computed from logged actions and propensities, ranks this risk \emph{across}
logging environments --- guidance for designing logs and choosing estimator families,
since it is weak at ranking candidate policies \emph{within} the single fixed log a practitioner
holds, and its cut point does not transfer.

Second, the optimizer's curse is not fixed by cross-fitting the outcome nuisance. When
the rule is fit on the data used to evaluate it, cross-fitting the nuisance alone leaves
the reuse bias in place and makes it worse. Honest policy-level splitting avoids the reuse
by targeting the learning procedure's value --- a change of estimand, not a de-biasing of
the full-sample policy.

Third, propensity-estimation error is the largest degradation we measure. Replacing the
exact propensity with an out-of-fold estimate hurts \est{IPS} more than any other stress
we apply, leaves doubly-robust estimation almost unchanged, and can invert the overlap
diagnostic itself.

Section~\ref{sec:cannot} states what this benchmark cannot tell you. Logging is
synthesized throughout and propensities are floored at $0.02$, so every failure we
document occurs with bounded weights; the floor also reduces the two tuned hybrids to
their untuned parents in most cells, leaving four practically distinct estimators. All
exact-value surfaces are synthetic or semi-synthetic --- hence the non-simulated check.
Section~\ref{sec:guide} distills the operational guidance. We release the benchmark;
public data only.
\end{abstract}

% ============================================================
\section{Introduction}
\label{sec:intro}

Consider a team allocating a fixed budget across a population: a retention offer to the
$20\%$ of customers most likely to respond, enrollment for the patients a treatment will
help most. Each is a \emph{budget-constrained allocation}: score every unit by an
estimated treatment effect and treat the highest-scoring until the budget is exhausted.
Before deploying a new rule the team wants to know what value it \emph{would} achieve,
from data already logged under the current one --- off-policy evaluation (OPE)
\citep{dudik2011doubly,thomas2016data}.

\looseness=-1 \textbf{The difficulty is exposure to weak overlap, not determinism per se.} The
deployable rule is a \emph{deterministic} top-$k$ policy: exactly one action per unit, so
$\pi_e$ puts probability $1$ on it. That does not by itself violate positivity --- the
relevant condition is $\pi_b(a_{\pi_e}(x)\mid x)>0$ for the actions $\pi_e$ selects. What
determinism removes is all averaging over actions: weights become
$\pi_e(a\mid x)/\pi_b(a\mid x)\in\{0,\,1/\pi_b(a\mid x)\}$, so wherever the logger rarely
took the action $\pi_e$ demands the weight is extreme and the effective sample collapses;
wherever $\pi_b$ has structural zeros, support genuinely fails. We exclude the second case
by construction --- propensities are floored at $0.02$, so every failure we document
occurs with weights bounded by $50$, arguably the more alarming finding. Practitioners
feel this as estimates wildly optimistic on some campaigns and fine on others, with little
guidance on which case they are in.

\looseness=-1 \textbf{We treat this as an empirical, decision-facing question.} Rather than propose a
new estimator, we ask: \emph{when} is offline evaluation of equal-cost top-$k$ allocation
trustworthy, \emph{which} estimator should a practitioner reach for, and \emph{can the
danger be detected from logged data alone}? Every estimate is paired with an explicit
reference value --- \emph{exact} ground truth where potential outcomes are known, a noisy
known-propensity Horvitz--Thompson \emph{reference estimate} on randomized trials --- and
the two categories are kept separate throughout. Existing work separately establishes
weak-overlap instability, doubly-robust estimation, ESS diagnostics, sample splitting and
policy-selection evaluation; our contribution is their integration under this estimand,
plus new empirical evidence.

We organize the results around four questions --- RQ1 estimator accuracy, RQ2 overlap
diagnostics, RQ3 the optimizer's curse under policy--evaluation reuse, RQ4 policy
selection --- and take RQ2 and RQ3 as primary, with the estimated-propensity boundary
(Contribution~3) scoping the first. Section~\ref{sec:guide} states the surviving claims
at the strength the evidence supports; a reader who wants conclusions without the
qualifications can read it first; Appendix~\ref{app:claims} tabulates every claim
against the evidence and scope supporting it, the fastest route to auditing
claim--evidence alignment.

\paragraph{Contributions.} We make three.
\begin{enumerate}[leftmargin=*,itemsep=2pt]
\item \textbf{A design correction: overlap in top-$k$ allocation is governed by
  logger--target misalignment, not logging sharpness alone (RQ2).}
  The argument is analytic first, empirical second. For a deterministic target the weight
  is $1/\pi_b(a_{\pi_e}(x)\mid x)$, so support is governed by the logger's probability of
  the target's \emph{actions}, not by how sharply it logs. Appendix~\ref{app:mechanics}
  turns that into the exact sharpening limit of Proposition~1
  (Section~\ref{sec:design}): a score-aligned logger's flat overlap is a
  finite-range plateau, while an action-aligned one sharpens toward full support. That is
  a prediction, and we tested it --- re-centring the logger at the budget cutoff produces
  the predicted reversal --- sharpening now \emph{improves} overlap, and the failure
  gradient sharpens to $0.0\%$/$1.7\%$/$26.7\%$ (Appendix~\ref{app:cutoff}).

  We offer this as a correction to how such benchmarks are built, not as a discovery: our
  own first design swept temperature on a score-aligned logger, never varying the quantity
  it meant to study. The empirical content is magnitude. Temperature alone
  moves median ESS only $0.562 \to 0.522$; crossing alignment with temperature moves it
  $0.56$ / $0.42$ / $0.19$, with \est{IPS} failure rates of $8\%$ / $13\%$ / $32\%$ at the
  sharpest temperature. ESS ranks that error across logging environments, though only weakly within a
  single log (ROC-AUC $0.85$ in-sample; $0.83$ and $0.91$ on two held-out families,
  the $0.91$ at a $2\%$ error target chosen for that suite's error scale, not selected
  for AUC; Table~\ref{tab:delta-grid}).

\looseness=-1   One caveat here: the design and ranking claims are not independent, since misalignment
  depresses coverage, a factor of ESS. The estimated-propensity boundary is
  Contribution~3.
\item \textbf{An estimand-coherent comparison of nuisance-only and policy-level splitting
  under policy--evaluation reuse (RQ3).} Cross-fitting only the outcome nuisance does not
  remove reuse bias --- frozen-policy cross-fitting is \emph{more} optimistic than plain
  DR. Honest policy-level splitting evaluates the learning procedure on independent folds,
  with bias magnitude $58$--$92\%$ smaller over eight known-effect regimes, seven of which
  share the \dataset{IHDP} covariate matrix (response surfaces, not populations). Honesty and
  sample splitting are long established
  \citep{athey2016recursive,chernozhukov2018double,athey2021policy}; the contribution is
  the estimand-coherent demonstration in the allocation-OPE setting.
\item \textbf{An empirical boundary for the diagnostic program: propensity estimation
  dominates, and can invert the screen (RQ2, continued).} Replacing the exact $\pi_b$
  with an out-of-fold estimate is the largest degradation we measure --- \est{IPS}
  failure rises from $6.3\%$ to $37$--$63\%$ of cells, dwarfing the $2.8$--$11.7\%$
  from moving the logger regime --- and a poor propensity model does not merely weaken
  the ESS screen but \emph{inverts} it (AUC $0.85$ to a coin flip, then $0.05$). The
  screen's prerequisite is a credible propensity model; Section~\ref{sec:propensity}
  gives the mechanism.
\end{enumerate}
\paragraph{What this paper establishes.} Stated affirmatively: overlap risk in top-$k$
allocation is governed by logger--target misalignment and is rankable ex ante, across
logging environments rather than within one log, from logged
data, \emph{conditional on a credible propensity model}; under policy--evaluation reuse,
nuisance-only cross-fitting is counterproductive, and honest policy-level splitting
evaluates the learning procedure without reuse, at the price of targeting a different
estimand; and doubly-robust estimation is the most
stable estimator across the \emph{individual} stresses we apply, hence a defensible
default when at least one nuisance model is credible. Three further findings are logically independent. The estimator
rankings (RQ1), the negative model-adequacy-screen result (Section~\ref{sec:omquality}),
and the policy-selection caution (RQ4) --- candidate-aligned logs
compare policy--logger pairs, not policies --- all matter to anyone building an OPE
selection benchmark. Supporting all of it is a released benchmark: six fixed-policy estimators over
five datasets, plus two $2{,}160$-configuration known-effect hardening sweeps.

\section{Related Work}
\label{sec:related}

\paragraph{OPE estimators and weak overlap.} The standard toolkit spans the direct method,
inverse propensity scoring and its self-normalized variant \citep{swaminathan2015self},
doubly-robust estimation \citep{dudik2011doubly}, and variance-controlled hybrids such as
Switch-DR \citep{wang2017optimal} and shrinkage-based DR \citep{su2020doubly}; formal
definitions are in Appendix~\ref{app:estimators}. Their behaviour is well characterized for
\emph{stochastic} targets, and effective sample size is an established weak-overlap
diagnostic \citep{austin2011introduction}. Deterministic and constrained targets
have received less benchmark attention \citep{guo2022offpolicy,nishimura2024coupon,
yeom2025breaking}, which is the gap this paper addresses. The closest neighbour is \citet{tanaka2026ope}, where determinism sits on the
\emph{opposite} side: their \emph{logging} policy is deterministic, so $\pi_b$ degenerates
and weighting is unidentified rather than merely noisy, and they recover randomness from
user-click stochasticity for slate rankings. Our determinism is in the \emph{target}, with
$\pi_b$ stochastic and floored, so positivity holds and the difficulty is variance, not
identification --- complementary problems, and their \emph{real} logged feedback is what
our synthesized logging lacks (Section~\ref{sec:cannot}). On the design side,
\citet{douglas2026logging} \emph{choose} the logging policy, trading reward against
coverage of the target's actions --- the design-side mirror of our fixed-log alignment
axis. Weight-reliability constraints appear inside off-policy \emph{learning} too
\citep{liu2022offline}; our contribution is validating ESS as an ex-ante risk ranking,
not the diagnostic itself.

\paragraph{Cross-fitting, honesty, and the optimizer's curse.} When a policy is fit and
evaluated on the same sample, its in-sample value is optimistically biased
\citep{smith2006optimizer}. Cross-fitting removes nuisance overfitting in DML
\citep{chernozhukov2018double}, and honesty --- evaluating the learning \emph{procedure}
on data not used to fit it --- is standard in the causal-forest and policy-learning
literature \citep{athey2016recursive,athey2021policy}. Our contribution is not that
principle but an estimand-coherent empirical demonstration in this setting, separating
nuisance-only cross-fitting (frozen policy) from honest policy-level splitting
(fold-specific policies, each scored against its own fold-matched reference); the former
does not address the observed reuse bias --- indeed it is \emph{more} optimistic than
plain DR (Section~\ref{sec:rq3}).

\looseness=-1 \paragraph{Policy selection and benchmarks.} OPE is used to \emph{select} among
candidates, and dedicated metrics quantify selection quality
\citep{saito2021counterfactual,kiyohara2024scope}. Semi-synthetic causal benchmarks
supply known-effect data \citep{hill2011bayesian,dorie2019automated}, and open bandit
datasets supply real logged feedback \citep{saito2021open}. In uplift modelling, \citet{li2026upliftbench} shows within-experiment ranking
metrics can invert the ordering ground-truth effect accuracy or policy risk would give
--- the selection question upstream of ours: there, which scoring rule to pick; here,
whether the induced fixed policy's value can be trusted from logged data. We combine both traditions, pair every estimate with an explicit reference value, and ---
as Section~\ref{sec:rq4} shows --- find that giving each candidate its own aligned log
confounds selection by comparing policy--logger \emph{pairs} rather than policies.

\section{Problem Setup}
\label{sec:setup}

\paragraph{Allocation value.} Each unit $i$ has context $x_i$, a binary action
$a\in\{0,1\}$ (withhold / treat), potential outcomes $Y_i(0),Y_i(1)$, and unit cost
$c_i$. A scoring rule $s(\cdot)$ and budget $k\in[0,1]$ induce a deterministic
allocation $a_{\pi}(x)$: treat units in descending score order while cumulative cost
stays within $k\sum_j c_j$ (with unit costs, ``treat the top-$k$ fraction''). The
estimand is the gross expected outcome under the allocation,
\begin{equation}
  \V \;=\; \mathbb{E}\big[\,Y\big(a_{\pi_e}(X)\big)\,\big],
  \label{eq:estimand}
\end{equation}
a value level (cost enters only through the budget constraint).
A top-$k$ rule is a \emph{batch} policy: whether unit $i$ is treated depends on the other
units through the empirical cutoff. We therefore target the \emph{cohort-conditional}
value $\V_n(z)=\frac1n\sum_i\mathbb E[Y_i(z_i)\mid X_{1:n}]$ for the fixed allocation
$z=a_{\pi_e}(X_{1:n})$ --- what our estimators and references compute --- not a
population-quantile policy with a predetermined cutoff. Uncertainty is correspondingly
conditional: the bootstrap holds allocation and cutoff fixed.

\looseness=-1 \textbf{Scope of the empirical study.} The formalism above admits heterogeneous costs
$c_i$, but \emph{every experiment in this paper sets $c_i\!=\!1$}, so the allocation is
always ``treat the top-$k$ fraction'' under a single binary action. We therefore study
\emph{equal-cost top-$k$ allocation}, not the general knapsack problem: we evaluate no
heterogeneous-cost or multi-action setting, and all logged feedback is synthesized
(Section~\ref{sec:design}) rather than drawn from a naturally occurring observational
logging policy. Claims and practitioner guidance below should be read at that scope.

\paragraph{Reference values.} Two explicitly separate categories. On data with known
potential-outcome means (synthetic, \dataset{IHDP}, the ACIC-style DGPs) we compute the
\emph{exact} value of the allocation. On randomized trials with known constant propensity
(\dataset{Hillstrom}, \dataset{Lenta}, \dataset{Jobs}) no exact value exists, so we use an
unbiased but noisy known-propensity Horvitz--Thompson \emph{reference estimate} on the
randomized evaluation split shared with the estimators. Error against the second is \emph{not} estimator MSE: it
combines OPE error with reference-estimation variance. We therefore never pool the two
categories in a single claim, and report cross-dataset averages separately for each.

\paragraph{Why standard OPE strains.} Because $\pi_e$ is deterministic, the importance
weight $w_i=\pi_e(a_i\mid x_i)/\pi_b(a_i\mid x_i)$ is either $0$ or
$1/\pi_b(a_i\mid x_i)$, and the latter blows up wherever $\pi_b$ seldom chose $\pi_e$'s
action. The strain is governed by the \emph{overlap} between $\pi_e$ and $\pi_b$, which we
control explicitly and measure.

\section{Benchmark Design}
\label{sec:design}

\paragraph{Datasets (Table~\ref{tab:datasets}, Appendix~\ref{app:estimators}).} Five datasets span three regimes.
\dataset{Synthetic} has known $\mu_0,\mu_1$ and a tunable effect, giving an exact
value and full control. \dataset{IHDP} (continuous outcome) and \dataset{Jobs}
(binary, randomized subset) are the canonical semi-synthetic causal benchmarks ---
\dataset{IHDP} supplies exact effects, \dataset{Jobs} a randomized reference.
\dataset{Hillstrom} and \dataset{Lenta} are real marketing RCTs with a known constant
treatment probability; \dataset{Hillstrom} is natively three-arm and we collapse its two
e-mail arms into one ``treat'' action ($\pi_1=2/3$), so it evaluates a coarsened action
space. Table~\ref{tab:datasets} reports sizes \emph{before} logging: rejection sampling
retains a subset, and on the small datasets that matters. \dataset{Jobs} falls to
$132$--$203$ logged units (Appendix~\ref{app:retained-n}), so it is noise-dominated and
the effective number of informative primary datasets is closer to four than five. That
caveat applies to every cross-dataset mean and five-cluster bootstrap here. Large
RCTs are capped at $50{,}000$ rows by uniform subsampling, preserving the constant
propensity.
As a known-effect \emph{hardening} layer, we add six ACIC-2017-style data-generating
processes over the real \dataset{IHDP} covariate matrix, following the competition's
``real covariates $+$ simulated response surfaces'' design \citep{hahn2019acic}: three
surface families (linear, nonlinear, step-subgroup effects) crossed with two noise
levels. Surface coefficients are fixed per setting (only the factual treatment and outcome
noise are redrawn per seed), noise is calibrated to the effect scale, and the baseline
surface carries a positive intercept so the gross-outcome estimand is bounded away from
zero at every budget --- avoiding the small-denominator artifact that inflates
relative-error metrics when $\V\!\approx\!0$.

\paragraph{Logging policies and the overlap knob.} For a randomized dataset we
synthesize logged feedback by \emph{rejection sampling}: a logging policy
$\pi_b(1\mid x)=\sigma\!\big(\tilde s(x)/\tau\big)$ is a temperature-$\tau$ softmax over
the \emph{standardized} candidate score $\tilde s = (s-\bar s)/\mathrm{sd}(s)$, and we
keep unit $i$ with probability proportional to
$\pi_b(a_i\mid x_i)/\pi_{\mathrm{rct}}(a_i\mid x_i)$; retained units keep their real
(action, outcome). The logging treat-probability is \emph{floored at $0.02$} for
numerical stability --- a design property with consequences worth stating up front:
importance weights are bounded by $50$ by construction, structural zeros never occur, and
the weak-overlap failure documented in Section~\ref{sec:results} is therefore a
\emph{bounded-weight} phenomenon, not an unbounded-tail one. We sweep
$\tau\in\{0.5,2.0,5.0\}$.

\textbf{Temperature alone is not a valid overlap parameter, and the reason is specific
to deterministic targets.} What matters is $\pi_b$ on the \emph{action the target
selects}. Our self-aligned logger matches the candidate's score \emph{ranking}, not its
action boundary: the logistic is centred at the score mean, while the top-$k$ boundary
sits at the budget cutoff. And indeed, over the tested range, sharpening barely moves
overlap. That is a finite-range plateau, not a guarantee --- extreme sharpening collapses
overlap through the low-probability band between mean and cutoff, while a genuinely
\emph{action-aligned} logger sharpens toward full support. The limit is exact:

\medskip
\noindent\textbf{Proposition 1 (sharpening limit).}
\emph{For a top-$k$ target under score $s$ and a logistic logger floored at
$\varepsilon$, the normalized Kish ESS satisfies
$\mathrm{ESS}/n \to \{\mathbb{E}[1/q_\tau(X)]\}^{-1}$ in the large-sample limit,
where $q_\tau(X)$ is the logging probability of the target's action. Consequently, for
a score-aligned logger, as $\tau \to 0$ this population ESS fraction satisfies}
\[
\frac{\mathrm{ESS}}{n} \;\longrightarrow\;
\Big[\frac{c}{1-\varepsilon}+\frac{1-c}{\varepsilon}\Big]^{-1}
\;\approx\; \frac{\varepsilon}{1-c},
\]
\emph{where $c = 1-\lvert F_{\tilde s}(\tilde q)-F_{\tilde s}(0)\rvert$ is the mass on
which logger and target agree in the limit ($\tilde q$ the standardized cutoff,
reducing to $c = k+F_{\tilde s}(0)$ for $\tilde q>0$); an
action-aligned logger (the same logistic centred at the top-$k$ cutoff) has an empty
mismatch band and instead sharpens toward full support, $\mathrm{ESS}/n \to
1-\varepsilon$. Derivation in Appendix~\ref{app:mechanics}.}
\medskip

For the benchmark's score distributions at $k{=}0.1$, $c\approx0.6$, so the
score-aligned limit is $\approx 0.05$. Extending
the grid downward confirms both branches on the benchmark's own data: at $k{=}0.1$,
$\tau{=}0.05$, median ESS is $0.061$ score-aligned but $0.960$ action-aligned
(Appendix~\ref{app:mechanics}).
Overlap is therefore governed by logger--target \emph{action} alignment, an explicit
experimental axis crossed with $\tau$ (Section~\ref{sec:rq2}). \dataset{IHDP}, which has known potential-outcome means but no recorded propensity,
instead draws actions from $\pi_b$ and rewards from the known surface, so it retains every
unit at every temperature --- the sole constant-$n$ control for separating overlap from
sample-size loss (Appendix~\ref{app:retained-n}). Two properties documented in
Appendix~\ref{app:logging} matter downstream: the logistic is centred at the score
\emph{mean}, not the budget cutoff, so $\tau$ is comparable across learners but not across
budgets; and rejection sampling retains fewer units at low $\tau$.
Appendix~\ref{app:cutoff} re-runs the full sweep with the cutoff-centred
logger. We keep the mean-centred sweep primary deliberately: it spans the wider overlap
range that a risk-\emph{ranking} study needs. Section~\ref{sec:rq2} reports both
failure gradients. In the main sweep $s$
is the \emph{candidate's own} score (\emph{self-aligned} logging): the logger shares the candidate's score, not its
top-$k$ action boundary, and is not the target policy. This isolates the overlap axis;
Section~\ref{sec:rq2} makes alignment itself an axis.

\looseness=-1 \paragraph{Estimators, metrics, and protocol.} Through one interface we evaluate six
estimators that all target the fixed deterministic policy $\V$ and are therefore ranked
on a common estimand: \est{DM}, \est{IPS}, self-normalized \est{SNIPS}
\citep{swaminathan2015self}, doubly-robust \est{DR} \citep{dudik2011doubly},
\est{Switch-DR} \citep{wang2017optimal}, and \est{mIPS}, an IPS variant whose propensity is mixed with a uniform-exploration
component (retained for completeness; its median relative RMSE lands within $0.001$ of
\est{IPS} on every dataset; both tuned hybrids are retained because their data-driven
tuning collapsing to the untuned parents is itself a finding, Appendix~\ref{app:mechanics}).
A perturbation-smoothed DR targets a \emph{different}, smoothed-policy estimand; it is
reported separately, scored against its own matched reference
(Appendix~\ref{sec:smoothing}). All model-based estimators share one LightGBM
\citep{ke2017lightgbm} outcome model per cell, and the two hybrids' hyper-parameters
(Switch-DR's threshold, mIPS's mixing weight) are tuned on the logged sample by an
estimated-MSE proxy rather than hard-coded, so no estimator gets a hand-picked advantage.
All estimators are our own implementations behind one interface (so nuisances are genuinely
shared); \est{IPS}, \est{SNIPS}, \est{DM} and \est{DR} agree with Open Bandit
Pipeline's reference implementations to floating-point precision on identical inputs
(\texttt{make reference-check}).
Because outcome scale spans $\sim\!30\times$ across datasets we report \emph{relative}
RMSE, RMSE$/|V_{\mathrm{ref}}|$, per cell across seeds and aggregated by median (robust to
the heavy IPS tail). Each dataset is split $50/50$: scoring rules are fit on \emph{train
only}; logged feedback, the reference value and all estimators operate on the evaluation
split. Formal definitions, aggregation formulas, leakage guards and the anomaly validator
are in Appendix~\ref{app:estimators}.

We sweep ten seeds, six budgets $k\in\{0.1,0.2,0.3,0.5,0.7,1.0\}$, three temperatures
and three candidate policies (\est{T-learner}, \est{S-learner}, random baseline).
A \emph{cell} is a (dataset, candidate policy, budget, temperature) tuple --- the unit
over which relative RMSE is computed, aggregating its ten seeds --- giving $270$ cells
across the five datasets ($2{,}700$ configurations); paired comparisons and cluster
bootstraps resample the unit named at each use. The full-budget point $k\!=\!1$ is
degenerate for \emph{selection} and excluded from RQ4; it is retained for RQ1--RQ2, where
excluding it leaves the ordering unchanged. The optimizer's-curse study uses a separate
protocol with the policy fit \emph{in-sample}.

\paragraph{Roadmap.} Table~\ref{tab:roadmap} maps each sweep to the axis it varies and
the section that consumes it; the rows map onto the \texttt{make} targets of
Appendix~\ref{app:repro-targets}, runnable from the anonymized repository accompanying
this submission; full reproduction takes roughly $17$ hours on a $10$-core laptop
(Appendix~\ref{app:repro-targets}).\footnote{\url{https://github.com/binshuangli/allocation-ope-bench}.}

\begin{table}[h]\centering
\caption{The experimental surface at a glance. Each sweep varies the named axes with
everything else held fixed; Appendix~\ref{app:repro-targets} lists the exact targets.}
\label{tab:roadmap}
\small
\begin{tabular}{lll}
\toprule
sweep & what it varies & consumed in \\
\midrule
primary accuracy + alignment & logger regime $\times$ $\tau$ $\times$ budget & \S\ref{sec:rq1real}, \S\ref{sec:rq2} \\
cutoff-centred rerun & logger centring (cutoff vs.\ score mean) & \S\ref{sec:rq2}, App.~\ref{app:cutoff} \\
propensity estimation & known $\pi_b \to$ out-of-fold $\hat\pi_b$ & \S\ref{sec:propensity} \\
outcome-model ladder & $\hat\mu$ class (LightGBM $\to$ ridge) & \S\ref{sec:omquality} \\
out-of-fold nuisance & in-sample $\to$ honest $\hat\mu$ & \S\ref{sec:rq1real} \\
floor sensitivity & propensity floor $\varepsilon$ to $0.0002$ & \S\ref{sec:rq1real}, App.~\ref{app:mechanics} \\
known-effect hardening ($\times 2$) & DGPs over real covariates & \S\ref{sec:robustness} \\
optimizer's curse & policy--evaluation reuse; splitting level & \S\ref{sec:rq3} \\
selection & slate $\times$ logging design & \S\ref{sec:rq4} \\
\dataset{Twins} & non-simulated paired reference & \S\ref{sec:twins-main}, App.~\ref{app:twins} \\
\bottomrule
\end{tabular}
\end{table}

\FloatBarrier
\subsection{What this benchmark cannot tell you}
\label{sec:cannot}

Five limits bound every claim below; we state them together so the results sections can
report findings at their supported strength without relitigating scope.

\begin{itemize}[leftmargin=*,itemsep=0pt]
\item \textbf{Which loggers are hard to estimate.} Section~\ref{sec:propensity} measures
  what estimated propensities cost \est{IPS} --- the largest degradation we observe --- but our
  logger is a smooth function of a fitted uplift score, correspondingly hard to recover. A
  simpler real-world logger would be estimated far more accurately, so read the
  $37$--$63\%$ failure rates as the hard case. Propensities remain floored at $0.02$.
\item \textbf{Observational logs.} All logged feedback is synthesized, never observed from
  a deployed logger --- which buys exact propensities and a controlled alignment axis at
  the cost of realism, and is what \citet{tanaka2026ope} have that we do not. The
  control is not merely cheaper, though: the RQ2 claim requires \emph{varying}
  logger--target alignment while target, data and truth stay fixed, and an observational
  log supplies one point on that axis, not the axis. That one point would test whether
  the diagnostics, computed on a real log, rank its realized error consistently with the
  synthesized axis --- a spot check --- but could not test the alignment mechanism (no
  counterfactual logger exists for the same population) nor calibrate cut points (a
  single environment). Adding it is the
  concrete next step --- a naturally logged bandit dataset with recorded propensities
  \citep{saito2021open} --- but not a drop-in: its multi-action recommendation setting
  is not equal-cost binary top-$k$ allocation, so the adaptation changes the object of
  study and is work of its own.
\item \textbf{Populations, and whose surfaces they are.} Seven of eight known-effect
  regimes share the \dataset{IHDP} covariate matrix, so $58$--$92\%$ is variation in
  response surfaces, not populations; both hardening suites reuse calibration
  covariates, so transfer is held out at the DGP level only. The exact-value evidence therefore remains
  synthetic or semi-synthetic throughout; Appendix~\ref{app:twins} removes that reliance
  on simulated response surfaces with a reference read off recorded paired outcomes,
  where the RQ1 ordering (at a larger margin, $5.0\times$ against $2.0\times$), the RQ2
  alignment mechanism and the RQ3 cross-fitting sign all replicate while the calibrated
  cut points do not.
\item \textbf{Inference.} Five primary datasets --- one, \dataset{Jobs}, noise-dominated
  at $132$--$203$ units --- so no cluster interval carries much information, and all
  intervals are \emph{conditional}, understating uncertainty for model-based estimators:
  measured coverage on exact-value cells is $0.89$ for \est{DM} against the nominal
  $0.95$, and \est{IPS} falls to $0.89$ in the flagged regime (Appendix~\ref{app:coverage}).
\item \textbf{Scope.} Binary action, unit costs, top-$k$ fraction; gross value $\V$, whose
  relative RMSE partly rewards baseline predictability and so flatters \est{DM} --- on
  \dataset{Twins} it compresses an \est{IPS} error of $62\%$ of the achievable policy
  spread to a relative RMSE of $0.016$ (Appendix~\ref{app:twins}). Clipped IPS and
  shrinkage-DR are both evaluated (Appendix~\ref{app:mechanics}); perturbation-DR is
  scored only against its own smoothed-policy reference (Appendix~\ref{sec:smoothing}).
  The floor's no-op argument is specific to tail control (clipping, switching,
  shrinkage); the balancing-weights family is outside the evaluated set --- a stated
  boundary, not a finding.
\end{itemize}

\section{Results}
\label{sec:results}

What survives everything below, in two sentences: overlap risk is governed by
logger--target \emph{action} alignment and is rankable \emph{across} logging
environments given credible propensities --- not within one log; and under
policy--evaluation reuse, only policy-level honesty addresses the optimism ---
nuisance-only cross-fitting makes it \emph{worse}. The rest of this section states each
finding at the strength its evidence supports, organized around four questions: accuracy
(RQ1), overlap diagnostics (RQ2), the optimizer's curse (RQ3), policy selection (RQ4)
--- with the outcome-model-quality axis between RQ2 and RQ3 and further robustness
checks in Appendix~\ref{app:moved}.

\subsection{RQ1 --- Which estimators are accurate?}
\label{sec:rq1}

\FloatBarrier
\begin{table}[H]\centering
\caption{RQ1 estimator accuracy: median relative RMSE (RMSE$/|V_{\mathrm{ref}}|$, normalized by the configuration-level reference --- exact where known, else the HT estimate); lower is better, best per column in bold. The last two columns re-run the benchmark with a five-fold \emph{out-of-fold} $\hat\mu$ (\texttt{make repro-full-oof}), all else fixed; \est{IPS}, \est{SNIPS} and \est{mIPS} use no outcome model and are identical across the two, so the change isolates nuisance honesty: the \est{DM}/\est{DR} ordering is unchanged, the model-based advantage on exact-value data persists, and on the RCT-reference column all six fall within $0.003$, which we do not read as a ranking. Exact-value and HT-reference averages are reported \emph{separately}, not pooled: they measure error against different references. Column blocks, left to right: per-dataset (in-sample $\hat\mu$), then cross-dataset means in two blocks --- in-sample and out-of-fold --- each split exact-value / HT-reference. Excluding the noise-dominated \dataset{Jobs}, the in-sample HT-reference average separates by only $0.002$ (\est{DM} $0.024$, \est{IPS} $0.026$; Section~\ref{sec:rq1real}).}
\label{tab:rq1-accuracy}
\small
\begin{tabular}{lrrrrrrrrr}
\toprule
& \multicolumn{5}{c}{per dataset (in-sample $\hat\mu$)} & \multicolumn{2}{c}{mean, in-sample} & \multicolumn{2}{c}{mean, out-of-fold} \\
\cmidrule(lr){2-6}\cmidrule(lr){7-8}\cmidrule(lr){9-10}
estimator & hillstrom & ihdp & jobs & lenta & synthetic & exact & RCT-ref & exact & RCT-ref \\
\midrule
dm & \textbf{0.022} & \textbf{0.014} & 0.066 & \textbf{0.026} & \textbf{0.044} & \textbf{0.029} & 0.038 & \textbf{0.030} & 0.041 \\
dr & 0.023 & 0.015 & 0.059 & 0.028 & 0.046 & 0.030 & \textbf{0.037} & 0.032 & 0.038 \\
switch\_dr & 0.023 & 0.015 & \textbf{0.058} & 0.028 & 0.046 & 0.030 & \textbf{0.037} & 0.032 & \textbf{0.038} \\
snips & \textbf{0.022} & 0.025 & 0.063 & 0.030 & 0.058 & 0.042 & 0.038 & 0.042 & 0.038 \\
ips & \textbf{0.022} & 0.052 & 0.068 & 0.030 & 0.062 & 0.057 & 0.040 & 0.057 & 0.040 \\
mIPS & \textbf{0.022} & 0.051 & 0.068 & 0.030 & 0.062 & 0.057 & 0.040 & 0.057 & 0.040 \\
\bottomrule
\end{tabular}
\end{table}

\begin{table}[H]\centering
\caption{The same in-sample errors under the \emph{incremental-value} renormalization, $|V(\pi_e)-V(\varnothing)|$, on the exact-value datasets (Section~\ref{sec:rq1real}). This is the metric that removes the baseline-predictability component gross value rewards: the family ordering persists, and the \est{DM}/\est{DR} gap closes.}
\label{tab:rq1-incremental}
\small
\begin{tabular}{lrr}
\toprule
estimator & ihdp & synthetic \\
\midrule
dm & \textbf{0.035} & \textbf{0.124} \\
dr & 0.039 & 0.125 \\
switch\_dr & 0.038 & 0.125 \\
snips & 0.065 & 0.201 \\
ips & 0.142 & 0.211 \\
mIPS & 0.142 & 0.215 \\
\bottomrule
\end{tabular}
\end{table}

\looseness=-1 Table~\ref{tab:rq1-accuracy} and Figure~\ref{fig:rq1} report median relative RMSE for the
six estimators targeting the fixed deterministic $V(\pi_e)$ (\est{perturbation-DR} targets
a smoothed-policy estimand; Appendix~\ref{sec:smoothing}). Model-based estimators lead on
four of five datasets and in aggregate; what matters is the \emph{magnitude} of that lead
and how much survives the checks below.

\looseness=-1 \textbf{The size of the gap depends on what we can compare against.} We report the
cross-dataset average separately by reference type, not pooled: the two measure error
against different objects. On the \emph{exact-value} datasets, where the
comparison is against true potential outcomes, the gap is large: \est{DM} $0.029$ and the
DR family $0.030$ versus $0.057$ for the IPS family, a factor of $\sim\!2$. On
the \emph{HT-reference} datasets the ordering compresses to near-nothing: the DR family
is nominally best ($0.037$) against \est{IPS}/\est{mIPS} $0.040$ and \est{DM} $0.038$,
a spread of $0.003$ we do not read as a ranking. A noisy reference adds a common error
floor that shrinks measurable differences; the next paragraph shows why. The gap
is widest on \dataset{IHDP}, where limited sample size ($n\!=\!672$), treatment
imbalance ($\pi_1\!=\!0.18$) and weak-overlap weights let a handful of observations
dominate the IPS average: relative RMSE $\approx\!0.052$, nearly $4\times$ that of
\est{DM} ($0.014$). \est{SNIPS} helps but does not close the gap.

\textbf{RCT-reference comparisons cannot reliably resolve estimator ordering.} Appendix~\ref{app:refdep} re-runs all three randomized datasets under a
matched design varying only whether the HT reference is computed on the \emph{same} units
the estimators see or on a \emph{disjoint} held-out third --- so the contrast isolates
reference dependence. The ordering moves with the reference. Under the shared one, a weighting or DR
estimator is best on all three datasets and \est{DM} on none. Under the disjoint one,
\est{DM} is best on all three, and \est{IPS} falls to last on \dataset{Jobs} ($0.1628$)
though not worst everywhere (\est{DR} on \dataset{Hillstrom}, \est{SNIPS} on
\dataset{Lenta}). Every estimator's error also rises by $48$--$114\%$ --- several times
the spread across estimators, so the reference supplies most of the measured error. That reference is itself an
inverse-propensity construction, so on shared units it shares sampling noise with
\est{IPS} and the two agree for reasons unrelated to accuracy --- with three datasets and
within-column differences of $0.001$--$0.009$ on the two large RCTs, evidence consistent
with that mechanism
rather than an isolation of it. It is why we decline to read the \est{IPS} column of
Table~\ref{tab:rq1-accuracy} as a ranking.

\dataset{Jobs} amplifies it: at $132$--$203$ retained units all six estimators land within
$0.058$--$0.068$, and excluding it the HT-reference average separates by only $0.002$
(\est{DM} $0.024$, \est{IPS} $0.026$) --- a direction, not a ranking.

\phantomsection\label{sec:rq1real}

\paragraph{Which differences are real?} Configurations share data, so we test the ordering
with a paired comparison: the per-configuration signed difference in relative RMSE between
two estimators, bootstrapped over dataset clusters. \est{IPS} is worse than \est{DR} by
$+0.015$ over all budgets (95\% CI $[+0.005,+0.027]$; $+0.013$ on the $k<1$ grid), and
worse in $72.6\%$ of cells. \est{Switch-DR} is statistically indistinguishable from
\est{DR} ($-0.0002$).

Two cautions attach to that number. The interval is a five-cluster bootstrap, narrow
because the within-dataset gaps are similar rather than because five clusters carry much
information; the same applies to every cluster interval here. And $+0.015$ is a
\emph{mean} paired difference, against a median of $+0.007$ (what
Table~\ref{tab:rq1-accuracy} reports): the mean is larger because the right tail dominates
it, and that tail is the practically important part.

\looseness=-1 These are self-aligned-logger magnitudes; Section~\ref{sec:rq2} shows the \est{IPS}
penalty is several times larger once the logger disagrees with the target.

Table~\ref{tab:rq1-incremental} reinforces this by removing the
baseline component of the gross-outcome value: accuracy is renormalized by the
\emph{incremental} value magnitude $|V(\pi_e)-V(\varnothing)|$ (treat-none baseline) on
the exact-value datasets. The model-based-vs-weighting ordering persists: the \est{IPS}
family sits at $0.142$--$0.215$ against $0.035$--$0.125$ for \est{DM}/\est{DR}. But
\est{DM} and \est{DR} become indistinguishable on synthetic ($0.124$ vs.\ $0.125$),
confirming that part of \est{DM}'s gross-value edge reflects baseline predictability.
This removes a reason to prefer \est{DM} on that renormalization; it is not itself an
argument for \est{DR}, whose case rests on Section~\ref{sec:omquality}.

\looseness=-1 \textbf{The family comparison does not depend on in-sample nuisance fitting.} Under a
five-fold \emph{out-of-fold} $\hat\mu$ the model-based family still beats \est{IPS} in
$89$--$100\%$ of the nuisance-cross-fitting check's $72$ configurations (four
exact-value DGPs only; $62$--$68\%$ over the whole benchmark's $225$, where the
HT-reference datasets compress every gap), and re-running the whole benchmark that way
(Table~\ref{tab:rq1-accuracy}; a controlled comparison, since \est{IPS}/\est{SNIPS}/\est{mIPS}
are numerically identical across blocks) leaves the family gap intact and the exact-value \est{DM}/\est{DR} ordering unchanged,
though \est{DM} becomes worst of six on \dataset{Jobs} and \dataset{Lenta}, and worst
of six on the HT-reference mean itself ($0.041$; Table~\ref{tab:rq1-oof}).

The practical read is that model-based estimators are more stable than pure
importance-weighting estimators \emph{when outcome models are reasonably predictive}
--- a condition that should be assessed with treatment-arm-specific held-out errors,
calibration checks, and sensitivity across outcome-model classes, not assumed; we
stress it directly in Section~\ref{sec:omquality}, where severe outcome-model
misspecification can invert even the within-family ordering on the nonlinear synthetic
DGP. Doubly-robust correction adds a safety margin; raw IPS is fragile precisely under
the deterministic target this paper studies.

\textbf{The propensity floor shapes the estimator set itself.} At
$\varepsilon{=}0.02$ importance weights are capped at $50$, and that ceiling --- not the
tuning --- is why the hybrids duplicate their parents: \est{Switch-DR} is bit-for-bit
identical to \est{DR} in $96.6\%$ of cells and \est{mIPS} reduces exactly to \est{IPS}
in $75.0\%$, while clipping is a no-op (Appendix~\ref{app:mechanics}). Sweeping the
floor itself to $\varepsilon{=}0.0002$ locates where that regime ends: tail control
begins to matter once weights exceed the $50$ ceiling (median \est{IPS} $0.111$ against
clipped $0.090$ at $\varepsilon{=}0.005$) while \est{DR} stays flat
(Appendix~\ref{app:robustness}). We keep $\varepsilon{=}0.02$ primary deliberately:
failures under bounded weights cannot be attributed to a few extreme weights (the more
alarming case, as Section~\ref{sec:intro} argues), and the sweep bounds what changes
below it. ``Clipping does not help'' is a property of the
bounded-weight regime we chose, not of allocation OPE.

\subsection{RQ2 --- Can you tell, in advance, when to trust OPE?}

\label{sec:rq2}

Accuracy is not constant: it is governed by overlap, overlap is governed by how much the
logger disagrees with the policy being evaluated, and both are observable without ground
truth.

\begin{figure}[t]
  \centering
  \includegraphics[width=0.95\linewidth]{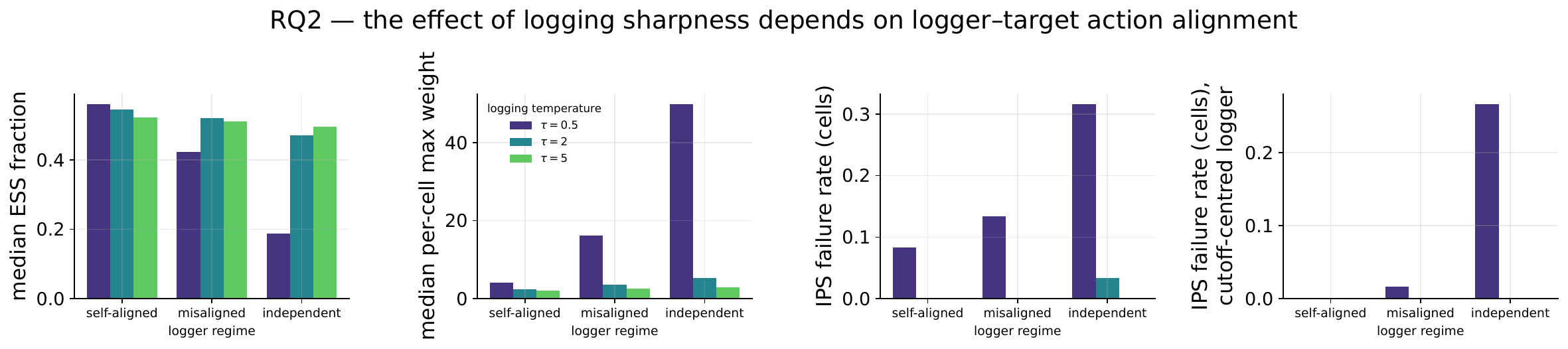}
  \caption{RQ2 --- overlap risk is a property of logger--target \emph{misalignment}, not
  of logging sharpness alone. Three logger regimes crossed with $\tau$, target fixed.
  Sharpening a \emph{self-aligned} (score-aligned) logger barely moves overlap over this
  range (left) --- a finite-range plateau, not a guarantee (Appendix~\ref{app:mechanics});
  once logger actions disagree with the target's, low $\tau$ collapses the effective
  sample, drives the maximum weight to the $0.02$-floor ceiling (centre), and raises the
  IPS failure rate (cells with median-over-seeds $|$relative bias$|>10\%$) from $8\%$ to
  $32\%$ (third panel). The fourth panel shows the same failure rates under
  budget-cutoff centring --- $0.0\%$, $1.7\%$ and $26.7\%$ at $\tau{=}0.5$, with
  failure at $\tau\geq2$ near zero (Appendix~\ref{app:cutoff}):
  the self-aligned failures follow from score-mean centring, and the misalignment
  gradient survives the correction.}
  \label{fig:rq2-alignment}
\end{figure}

\textbf{The overlap axis is alignment, not temperature.} We cross three logger regimes
with $\tau\in\{0.5,2.0,5.0\}$: \emph{self-aligned}, built from the evaluated candidate's
score but not its top-$k$ boundary (the main sweep's design); \emph{misaligned}, built
from the other learned candidate, a plausible incumbent; and \emph{independent}, built
from a candidate-independent score. Seeds, splits, fitted candidates and truth labels are
held fixed, so differences are attributable to the logger alone ($540$ cells). Retained
$n$ is not a confound --- the regimes retain near-identical medians ($1{,}793$ / $1{,}793$
/ $1{,}800$ at $\tau{=}0.5$ on \dataset{synthetic}), and \dataset{IHDP} holds $n{=}336$ in
all three.

\looseness=-1 Figure~\ref{fig:rq2-alignment} reports the result. At $\tau{=}0.5$ the median maximum
weight runs $4.0 \to 16.2 \to 50.0$ across the regimes, median ESS runs
$0.56 \to 0.42 \to 0.19$, and \est{IPS} failure rates run
$8.3\% \to 13.3\% \to 31.7\%$ ($2.8\% \to 4.4\% \to 11.7\%$ pooled,
Table~\ref{tab:rq2-conditioning}). Under the cutoff-centred logger the same
gradient is $0.0\%$ / $1.7\%$ / $26.7\%$ (Appendix~\ref{app:cutoff}; Section~\ref{sec:design}
states why the mean-centred sweep stays primary). The screen does not depend on that
choice: on the cutoff-centred sweep the association replicates, $\rho(\text{ESS})=-0.32$
pooled and $-0.58$ under the independent logger, against $-0.41$ and $-0.56$ on the
primary sweep --- the centring choice moves \emph{which} cells fail, not whether ESS
ranks failure. Within the self-aligned regime, temperature does
almost nothing. The practitioner's question is therefore not ``how smooth was my logger''
but ``how far is the policy I want to deploy from the one that collected my data''.

\looseness=-1 Can a practitioner detect this without ground truth? Across the $540$ cells the
logged-data diagnostics do rank realized \est{IPS} error:
$\rho(\text{ESS})=-0.41$, $\rho(\text{support deficiency})=+0.42$,
$\rho(\text{max weight})=+0.35$ --- smaller than a temperature-only design would report.

Cells share datasets, seeds and policies, so a cell-level $p$-value would be
anti-conservative. The trustworthy evidence is \emph{consistency}: within each dataset the
ESS correlation ranges $-0.35$ to $-0.71$ and support deficiency $+0.46$ to $+0.69$, and
leave-one-dataset-out values stay at $|\rho|=0.37$--$0.45$.

Appendix~\ref{app:rq2-cond-full} works through the
conditioning ladder in Table~\ref{tab:rq2-conditioning}: the ranking survives conditioning on dataset (negative in $5/5$,
$-0.35$ to $-0.71$) and on budget (AUC $0.68$--$0.95$ across strata), and survives
within the logger regime that actually produces failures ($\rho=-0.56$ under an
independent logger, $11.7\%$ of cells unsafe); but it \emph{collapses} under joint
regime-and-$\tau$ conditioning (median $-0.01$, negative in $6$ of $9$ strata). We read
that as expected rather than damaging --- regime $\times$ $\tau$ \emph{is} the mechanism
generating overlap variation here, so fixing both removes what ESS exists to rank --- but
it does narrow the claim: ESS ranks risk \emph{across} the logging situations a
practitioner might face, not \emph{within} a fixed one.

The diagnostics also generalize beyond the data they were calibrated on, which is the
practically useful test. A simple \emph{fragility screen} --- best understood as a
risk-\emph{ranking} heuristic, not a decision rule --- is fit on this benchmark (flag
a cell when ESS fraction $<0.36$ or support deficiency $>0.02$, targeting $|$relative
bias$|>10\%$). ROC-AUCs are for the \emph{continuous} ESS
fraction, not the binary rule (ESS ROC-AUC $0.85$ in-sample) and applied to the two held-out hardening
suites, run on the same alignment axis, whose DGPs were not used to calibrate it. On the
IHDP-covariate suite it flags $21\%$ of cells, with error rate $52\%$ among flagged
versus $7\%$ among unflagged (base rate $17\%$; ESS ROC-AUC $0.83$; precision $0.52$,
recall $0.67$). These quantities need only the logged actions and propensities, so the
indicator is computable before believing any estimate.

\looseness=-1 \textbf{The second suite reports a boundary rather than a failure.} On the
\dataset{Hillstrom}-covariate suite \emph{no} cell reaches $10\%$ relative bias, so the AUC
is undefined there. The cause is sample size, not a breakdown of the diagnostic: it retains
a median $5{,}000$ logged units against $336$ and $2{,}072$ elsewhere, with median
$|$relative bias$|$ $0.012$ against $0.056$ and $0.035$. At a target matched to its own
error scale the screen is at its strongest anywhere in the paper (ROC-AUC $0.91$ at
$\delta\!=\!2\%$, precision $0.58$, recall $0.84$); we report the full $\delta$ grid (Table~\ref{tab:delta-grid},
Appendix~\ref{app:rq2-cond-full}) rather than a per-suite best, since choosing $\delta$
post hoc would fit the held-out data. The
\emph{ranking} transfers while the \emph{cut point} does not --- recalibrate on your own
$n$. Appendix~\ref{app:cutoff} re-runs the benchmark under the
cutoff-centred logger.

\looseness=-1 \textbf{Within a single log the screen is much weaker.} Those correlations are
\emph{across} logging situations; a practitioner holds one. Fixing dataset, regime,
temperature and seed and ranking the $12$ candidate$\times$budget targets on that one log
($450$ logs) gives median $\rho=-0.11$, correct in sign in $58\%$ of logs and no better
than chance under the candidate-independent logger; it is negative only where overlap
varies within the log ($-0.30$ at $\tau{=}0.5$, $0.00$ at $\tau{=}5$). ESS ranks
\emph{which logging situation} carries risk, not which candidate to trust within one. \emph{Absolute} ESS ranks no better in practice ($0.858$ against the fraction's $0.851$; identical
on each suite, where constant $n$ makes them monotone transforms): what fails to transfer
is the mapping from diagnostic value to error level, not the statistic.

\subsection{RQ2, continued --- estimated propensities are the largest degradation, and
can invert the diagnostic}
\label{sec:propensity}

Everything above hands the weighting estimators the exact $\pi_b$; in real observational
logs propensity-estimation error is first-order. We re-run the alignment sweep with
$\pi_b$ replaced by an out-of-fold $\hat\pi_b$ (\est{LightGBM} on $x$, logistic on $x$, or
the sample treat rate), floored at the same $0.02$ (Table~\ref{tab:propensity}). Weights
\emph{and} diagnostics use $\hat\pi_b$: a screen that works only on propensities you do
not know would be useless.

\textbf{For \est{IPS}, estimation error dominates everything else measured here.} The
share of cells in which \est{IPS} exceeds $10\%$ relative bias rises from $6.3\%$ to
$37$--$63\%$ --- against $2.8\%$ to $11.7\%$ for moving the logger regime itself, on the
same cell-level unit. Our logger is a smooth function of a fitted uplift score; the best
$\hat\pi_b$ correlates only $\rho=0.30$ with the truth (mean absolute error $0.156$;
the logistic and marginal reach $0.126$/$0.129$, but only the marginal loses the rank
information, $\rho=-0.02$ against $+0.27$). The degradation is \emph{not} monotone in model quality. The flexible $\hat\pi_b$
damages \est{IPS} most (median per-cell max weight $11.8$ against $2.1$ for the marginal
model), trading bias for variance that \est{IPS} cannot absorb. The \emph{screen}, by
contrast, degrades monotonically, because what it loses is dispersion information. The
table's two \est{IPS} columns therefore point opposite ways, and the logistic model is
least damaging by median error while still failing half its cells.
\est{DM} is unchanged by construction (the pipeline check), \est{DR} and \est{Switch-DR}
are essentially immune ($0.037 \to 0.036$--$0.042$), and \est{IPS} degrades
$1.9$--$4.9\times$: the outcome-model term absorbs what the propensity model gets wrong.

\textbf{The screen is only as good as the propensity model, and can invert.} Under a
flexible $\hat\pi_b$ it is undamaged (ESS ROC-AUC $0.84$, matching the $0.85$ it reaches
with the exact propensity); under a logistic model it is a coin flip ($0.54$); under the
marginal model it is \emph{actively misleading} (AUC $0.05$, median \est{IPS} error
rising from $0.037$ in the lowest estimated-ESS quartile to $0.349$ in the highest). The mechanism is specific: ESS factors as coverage $\times$ dispersion, and a
near-constant $\hat\pi_b$ carries no dispersion, so $\widehat{\mathrm{ESS}}$ collapses to
coverage --- which is highest exactly where the true propensity is most extreme. The
inversion is systematic, not noise (Appendix~\ref{app:propensity-mech}). The propensity
model is a \emph{prerequisite}, not a caveat.

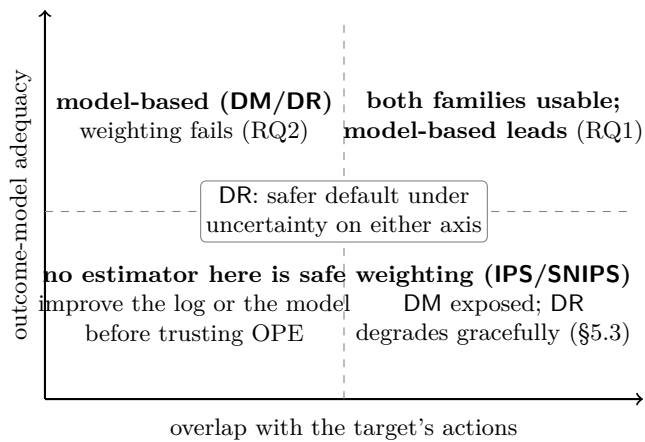
\begin{figure}[H]
\centering
\begin{tikzpicture}[scale=0.92, font=\small]
  \draw[->, thick] (0,0) -- (8.6,0);
  \node[below] at (4.3,-0.15) {overlap with the target's actions};
  \draw[->, thick] (0,0) -- (0,5.6);
  \node[rotate=90] at (-0.35,2.8) {outcome-model adequacy};
  \draw[dashed, gray] (4.3,0) -- (4.3,5.4);
  \draw[dashed, gray] (0,2.7) -- (8.4,2.7);
  \node[align=center] at (2.15,4.05) {\textbf{model-based (\est{DM}/\est{DR})}\\weighting fails (RQ2)};
  \node[align=center] at (6.45,4.05) {\textbf{both families usable;}\\\textbf{model-based leads} (RQ1)};
  \node[align=center] at (2.15,1.35) {\textbf{no estimator here is safe}\\improve the log or the model\\before trusting OPE};
  \node[align=center] at (6.45,1.35) {\textbf{weighting (\est{IPS}/\est{SNIPS})}\\\est{DM} exposed; \est{DR}\\degrades gracefully (\S\ref{sec:omquality})};
  \node[align=center, fill=white, inner sep=2pt, draw=gray, rounded corners=2pt] at (4.3,2.7) {\est{DR}: safer default under\\uncertainty on either axis};
\end{tikzpicture}
\caption{Schematic: the two-dimensional decision map of
Sections~\ref{sec:rq1real}--\ref{sec:omquality}, drawn. Positions are qualitative, not
calibrated coordinates. The overlap axis is rankable \emph{across} logging
environments by $\widehat{\mathrm{ESS}}$ given credible propensities
(Section~\ref{sec:rq2}) --- not within one log, and the cut points do not transfer. The
adequacy axis has \emph{no} transferable screen (the negative result of
Section~\ref{sec:omquality}); judge it by held-out error per arm.}
\label{fig:decision-map}
\end{figure}

\paragraph{The second axis: outcome-model quality.}\label{sec:omquality}
The RQ1 lead of model-based estimators is conditional on an adequate outcome model, so we
degrade $\hat\mu$ along a ladder (LightGBM $\to$ stumps $\to$ ridge). The effect is
DGP-specific and, on the nonlinear \dataset{synthetic} surface, decisive: degrading
$\hat\mu$ makes \est{DM} \emph{worst} of the four compared ($0.036 \to 0.099$, overtaking
\est{IPS} at $0.058$) while \est{DR} degrades gracefully ($0.037 \to 0.049$) and never
becomes worst. The crossing is beyond seed noise: seed-bootstrap intervals put the
\est{DM}$-$\est{IPS} gap at the stump rung at $[+0.022,\,+0.063]$ --- \est{DM} worse in
$100\%$ of $2{,}000$ seed resamples --- while \est{DR}'s degradation interval,
$[-0.001,\,+0.019]$, is consistent with graceful. On the other two DGP families
\est{DM} stays best or within a whisker at
every rung, so the inversion is a property of one family out of three.
The practical map is therefore two-dimensional --- overlap bounds weighting estimators,
outcome-model quality bounds model-based ones --- and \est{DR} is the safer default when
adequacy is uncertain. Figure~\ref{fig:decision-map} draws it.

 We also report a \emph{negative} result: the obvious analogue of
the ESS screen for model fragility, out-of-fold factual RMSE of $\hat\mu$, does \emph{not}
transfer ($\rho\!\approx\!-0.10$, AUC $0.48$), because factual predictive error conflates
irreducible noise with structural misfit. Full grid in Appendix~\ref{app:robustness}.

\subsection{RQ3 --- The optimizer's curse: what addresses policy--evaluation reuse?}

\label{sec:rq3}

When the candidate policy is \emph{learned on the same data used to evaluate it}, the
optimizer selects units whose scores are inflated by their own estimation noise, and plain
DR overstates the deployed value \citep{smith2006optimizer}. Fair comparison requires care
with \emph{estimands}. We compare two fixed-policy estimators --- plain \est{DR} and a
\emph{frozen-policy} cross-fitted DR (in-sample policy fixed; only the nuisance fit
out-of-fold) --- each scored against the true value of that same in-sample-optimized
policy. Separately we evaluate the \emph{honest fold-policy pipeline}: the policy is
re-learned per fold on out-of-fold data and evaluated on the held-out fold, so its
estimand is the value of the learning \emph{algorithm}, scored against its own
fold-matched exact reference. Mixing the two would conflate curse removal with the
estimand gap, so we do not; ``true value'' is \emph{exact} on the datasets carrying the
result. (\est{Perturbation-DR} targets a third estimand; Appendix~\ref{sec:smoothing}.)

\begin{table}[H]\centering
\caption{RQ3 optimizer's curse, per dataset: plain-DR $|$bias$|$ (in-sample policy value); the $|$bias$|$ change from nuisance-only cross-fitting (same fixed-policy estimand); and the bias reduction under honest \emph{algorithm} evaluation (a different estimand --- the learning procedure's value). ``Material optimistic bias'' is an explicit criterion --- plain DR's mean \emph{signed} bias exceeds $2\%$ of $|V_{\mathrm{ref}}|$ (third column) --- and signed on purpose: \dataset{Jobs} ($-0.2\%$) and \dataset{Lenta} ($-3.8\%$) carry non-trivial $|$bias$|$ but no optimism, so there is no upward reuse bias to remove; \dataset{IHDP} ($+6.9\%$) and \dataset{synthetic} ($+23.6\%$) are flagged. DR $|$bias$|$ is shown raw and as a fraction of $|V_{\mathrm{ref}}|$, since raw magnitudes are not comparable across outcome scales. Italic entries are out of scope --- on those datasets there is no material upward reuse bias to remove, so a ``\% removed'' is not comparable to the flagged rows; they are shown for completeness and nothing in the text depends on them. \dataset{Jobs}' $+8.4$ is the only cell in the benchmark where nuisance-only cross-fitting helps, on the noise-dominated dataset.}
\label{tab:rq3-debias}
\small
\begin{tabular}{lcrrrrr}
\toprule
dataset & material optimistic & DR bias$/|V|$ & DR $|$bias$|$ & DR $|$bias$|/|V|$ & nuis.\ x-fit & honest-algo. \\
 & bias? & (signed) & & & \% removed & \% removed \\
\midrule
hillstrom & no & -0.000 & 0.001 & 0.005 & {\footnotesize\itshape $-19.0$} & {\footnotesize\itshape 11.2} \\
ihdp & yes & 0.069 & 0.242 & 0.071 & -18.3 & 68.5 \\
jobs & no & -0.002 & 0.034 & 0.041 & {\footnotesize\itshape $+8.4$} & {\footnotesize\itshape 23.6} \\
lenta & no & -0.038 & 0.007 & 0.038 & {\footnotesize\itshape $-138.6$} & {\footnotesize\itshape 73.3} \\
synthetic & yes & 0.236 & 0.181 & 0.236 & -36.0 & 91.7 \\
\midrule
\multicolumn{7}{l}{\emph{six ACIC-style known-effect DGPs on real \dataset{IHDP} covariates (Section~\ref{sec:robustness}):}} \\
acic\_s1 & yes & 0.059 & 0.140 & 0.059 & -26.7 & 57.8 \\
acic\_s2 & yes & 0.394 & 0.858 & 0.394 & -16.0 & 77.7 \\
acic\_s3 & yes & 0.092 & 0.200 & 0.092 & -19.3 & 82.2 \\
acic\_s4 & yes & 0.381 & 0.891 & 0.381 & -16.5 & 79.7 \\
acic\_s5 & yes & 0.064 & 0.137 & 0.065 & -19.3 & 71.1 \\
acic\_s6 & yes & 0.145 & 0.303 & 0.145 & -17.1 & 77.3 \\
\bottomrule
\end{tabular}
\end{table}

\textbf{The central result: cross-fitting the outcome nuisance --- the reflexive fix ---
does not remove the reuse bias but makes it \emph{worse}.} With the policy frozen, an
out-of-fold $\hat\mu$ makes the DR correction \emph{honest}, and that is the problem: the
optimizer selected units whose scores were inflated by their own noise.

\emph{Observation (informal).} For a policy selected data-adaptively on the evaluation
sample, in the regimes we study the conditional residual is positive ---
$\mathbb{E}[y-\hat\mu \mid \text{selected}]>0$ under an honest $\hat\mu$ --- while an
in-sample $\hat\mu$ shrinks exactly that conditional residual. The DR correction
averages $w\,(y-\hat\mu)$, so honesty adds the optimism back. The difference between
the two pipelines is in fact exact algebra: by linearity of DR,
\[
\widehat V_{\text{in}} - \widehat V_{\text{honest}}
= \tfrac1n \textstyle\sum_i \big(1 - w_i\,\mathbb{1}[a_i{=}\pi_e(x_i)]\big)
\,\big(\hat\mu_{\text{in}} - \hat\mu_{\text{honest}}\big)(x_i, \pi_e(x_i)),
\]
whose expectation is $-\mathrm{Cov}\big(w\,\mathbb{1}[a{=}\pi_e],\,
\hat\mu_{\text{in}}\big)$ once cross-fitting makes $\hat\mu_{\text{honest}}$
independent of the evaluation draw \emph{and} the propensities are known, so
$\mathbb{E}[w\,\mathbb{1}[a{=}\pi_e] \mid x]=1$ --- as they are here by
construction. The sign condition is that the in-sample
nuisance be elevated exactly where the estimator's weighted observation indicator
is --- which own-observation fitting produces and cross-fitting removes. The identity
is exact; whether the covariance is positive and \emph{material} for a given learner
and logging regime is empirical, and that is what the eight regimes and the ablation
test. The frozen-policy cross-fit is therefore \emph{more} optimistic than plain DR
($+0.25$ / $+0.28$ vs.\ $+0.18$ / $+0.24$; ``\% removed'' negative, $-16\%$ to $-36\%$
across all eight known-effect regimes).

A decoupled ablation confirms the sign is not an artifact of the policy and nuisance
sharing a model: with an in-sample \est{LightGBM} policy and a different-class \est{DR}
nuisance (\texttt{make repro-optbias-decoupled}), cross-fitting stays more optimistic on
four of five datasets (pooled $-32\%$). The exception is \dataset{synthetic} at $+1\%$
--- essentially no effect either way --- where the decoupled ridge nuisance is the class
Section~\ref{sec:omquality} shows is misspecified on that nonlinear surface, so
model-class bias plausibly swamps the selection-noise channel there. Scope: \emph{for the data-adaptive top-$k$
learners studied here}, frozen-policy nuisance cross-fitting consistently increased
optimism across the eight primary known-effect regimes; in the decoupled ablation it
increased optimism on four of five datasets and was essentially neutral on the fifth. We
do not establish when the covariance is \emph{material} --- and the obvious toy model
illustrates why magnitude resists simple derivation: in a homogeneous normal-means
case with mild weights the covariance term is numerically negligible, leaving total
DR optimism essentially invariant to $\hat\mu$ there. The sign's form is the identity above; its
size in our regimes is measured, not derived. Seven of eight regimes share one
covariate population. The bias lives in the
policy--data dependence, not in nuisance reuse.

\looseness=-1 \textbf{What the practitioner gains from honest splitting.} Split at the \emph{policy}
level: re-learn the rule per fold on out-of-fold data, evaluate on the held-out fold. Against its own fold-matched reference that cuts $|$bias$|$ by $91.7\%$ and $68.5\%$ ---
close to definitional. The practitioner-facing question
is how much of the optimism they would otherwise carry is removed, measured against the
true value of the \emph{full-sample} in-sample policy --- what plain DR was estimating. On
that target the honest pipeline removes $89\%$ and $69\%$ of the per-cell error, and
\dataset{IHDP}'s $69\%$ nearly matches the $68.5\%$ above: the estimand gap costs
little. Seed-level bootstrap CIs $[88,95]\%$ and $[49,81]\%$.

\looseness=-1 \textbf{And the regime in which any of this matters.} Material bias --- plain DR's mean
signed bias above $2\%$ of $|V_{\mathrm{ref}}|$ --- appears only on the two datasets with
continuous, known effects where the in-sample model can overfit (\dataset{synthetic}
$+23.6\%$, \dataset{IHDP} $+6.9\%$), not on the three binary RCTs ($-0.2\%$, $-3.8\%$,
$-0.0\%$), whose non-trivial $|$bias$|$ is not \emph{optimistic}. ``\% removed'' uses the
conservative mean per-cell absolute bias throughout (Appendix~\ref{app:paired}).

\looseness=-1 \textbf{The pattern is not an artifact of two generators.} We re-run the same
comparison on the six ACIC-style exact-value DGPs (Section~\ref{sec:robustness}), which
also have known potential outcomes and an overfittable in-sample model. On all six,
plain DR is again optimistically biased (signed $+0.14$ to $+0.89$), nuisance-only
cross-fitting again \emph{worsens} it ($-16\%$ to $-27\%$), and honest policy-level
splitting again reduces the bias, by $58$--$82\%$ (per-setting values in
Table~\ref{tab:rq3-debias}; every seed-bootstrap interval excludes zero). With the two
original generators the reduction spans $58$--$92\%$ over all eight
continuous/known-effect regimes with uniform direction. That range is over \emph{point} estimates --- individual intervals are wider and extend
below its floor (\dataset{IHDP} $[49,81]\%$); Section~\ref{sec:cannot} bounds its breadth.

\looseness=-1 \textbf{The estimand distinction, stated once.} The $89\%$/$69\%$ figures above are an
estimand-gap diagnostic, not a bias measurement: against the full-sample target the honest
estimate sits at $+0.017$ and $-0.019$ versus plain DR's \emph{signed} $+0.181$ and
$+0.237$ (Table~\ref{tab:rq3-debias} reports mean per-cell absolute bias, $0.181$/$0.242$),
but the honest pipeline is not an unbiased estimator of that target. We are not de-biasing
a fixed data-adaptive quantity; we are honestly evaluating the policy-\emph{learning
procedure}, usually the deployment-relevant object anyway (Appendix~\ref{app:paired}
separates the signed and absolute readings).

\subsection{RQ4 --- Do estimators pick the right policy?}
\label{sec:rq4}

Finally: does OPE pick the best candidate under a budget? ``Best'' is the
value-maximizing candidate on the exact-value datasets and the HT-reference-best policy on
the RCT ones. We exclude the full-budget point, where every candidate treats everyone and
``best'' is an arbitrary tie.

\looseness=-1 \textbf{The logging design is a first-order confounder, so we vary it explicitly.} The
main sweep logs from \emph{each candidate's own} score. That is right for isolating the
overlap axis and wrong for selection: comparing $\Vhat_j$ across candidates then
compares policy--logger \emph{pairs} rather than policies, and real selection has one
historical log. We therefore re-run selection with a \emph{common logger} --- one logged
dataset per cell, identical for every candidate --- either aligned with the
\est{T-learner} or candidate-\emph{independent}. Table~\ref{tab:rq4-logger} reports all
three designs, with seeds, splits, candidates and truth labels fixed, on the
three-candidate slate of Section~\ref{sec:design} (baseline $1/3$) and a competitive
seven-policy slate (six candidates on the two continuous-outcome
exact-value datasets, seven on the three RCT ones; Appendix~\ref{app:rq4-percand}),
baseline $\approx0.15$ rather than $1/7$.

\textbf{In correct-selection accuracy the model-based edge survives every design;
regret is more qualified.} The $\est{DM}-\est{IPS}$ gap runs $+0.083 \to +0.105 \to +0.056$ across the per-candidate, common-aligned and common-independent designs on the
three-candidate slate, and $+0.044 \to +0.020 \to +0.016$ on the seven-policy slate. The
design therefore does \emph{not} dominate the estimator comparison: the gap moves by about
half its own size with no consistent direction, while a competitive slate shrinks it by
$1.9$--$5.3\times$ --- a null: no detectable logging-design effect, and a slate effect
larger than any we could resolve. With five datasets no single interval is authoritative,
so Table~\ref{tab:rq4-logger} layers a per-dataset sign count, a leave-one-dataset-out
range and two cluster bootstraps, all conditional on this benchmark. On the
three-candidate slate the finer bootstrap excludes zero under the common-aligned design,
sits \emph{at} zero to resampling resolution under the per-candidate one (its lower bound
changes sign across bootstrap seeds) and includes zero under the common-independent one,
while dataset-level clustering excludes it nowhere; on the competitive slate support is
absent at every level.

\textbf{Normalized regret agrees on the easy slate and dissents on the competitive one.}
Regret asks how costly the mistake was, not merely whether the top pick was right. On the
three-candidate slate the $\est{DM}-\est{IPS}$ regret gap favours \est{DM} under every
logging design ($-0.026$, $-0.052$, $-0.022$; lower is better). On the seven-policy slate
the sign reverses ($+0.045$, $+0.012$, $+0.016$): \est{IPS} incurs slightly \emph{lower}
regret. The summary is simple: on the easy slate both criteria favour \est{DM}; on the
competitive slate they disagree, and the disagreement tracks reference type. The $+0.045$
regret gap is not noise --- it is the size of the winner-accuracy gap on that slate
($+0.044$), pointing the other way --- and it splits exactly as winner accuracy does: the
two exact-value datasets favour \est{DM} ($-0.206$, $-0.018$), the RCT datasets favour
\est{IPS} (Appendix~\ref{app:rq4-perdataset}). The separation we can defend is therefore
the one on exact-value data, and neither criterion shows the gap \emph{closing} as the
logger becomes candidate-independent (Figure~\ref{fig:rq4}).

\begin{table}[!htbp]\centering
\caption{RQ4 selection quality by logging design: paired $\est{DM}-\est{IPS}$ gap,
per-dataset sign count, LODO range, two cluster bootstraps --- all \emph{conditional on
this benchmark} ($k{=}1$ excluded throughout RQ4). Dataset-level clustering excludes zero
nowhere; a competitive slate shrinks the gap and removes support at every level
(Section~\ref{sec:rq4}).}
\label{tab:rq4-logger}
\footnotesize
\begin{tabular}{llrrrrr}
\toprule
slate & logger & \est{DM} & \est{IPS} & gap & pos.\ / LODO & cluster CIs \\
 & & & & & & (ds$\times$seed; ds) \\
\midrule
\multirow{3}{*}{\shortstack[l]{3 cand.\\(base $0.33$)}}
 & per-candidate       & 0.656 & 0.573 & $+0.083$ & 4/5; $[.03,.14]$ & $[.00,.16]$; $[-.05,.21]$ \\
 & common, aligned     & 0.679 & 0.573 & $+0.105$ & 4/5; $[.06,.16]$ & $[.04,.17]$; $[-.02,.22]$ \\
 & common, independent & 0.661 & 0.605 & $+0.056$ & 3/5; $[-.02,.12]$ & $[-.03,.15]$; $[-.10,.23]$ \\
\midrule
\multirow{3}{*}{\shortstack[l]{7 cand.\\(base $\approx0.15$)}}
 & per-candidate       & 0.358 & 0.313 & $+0.044$ & 2/5; $[-.04,.12]$ & $[-.05,.14]$; $[-.16,.26]$ \\
 & common, aligned     & 0.344 & 0.324 & $+0.020$ & 3/5; $[-.07,.12]$ & $[-.09,.12]$; $[-.21,.25]$ \\
 & common, independent & 0.338 & 0.322 & $+0.016$ & 2/5; $[-.06,.09]$ & $[-.08,.12]$; $[-.20,.23]$ \\
\midrule
\bottomrule
\end{tabular}
\end{table}

\textbf{Two further readings, developed in Appendices~\ref{app:rq4-extra}
and~\ref{app:rq4-percand}.} Pooled rates hide a reference-type split: on exact-value data
the model-based advantage is large even under a shared independent log ($+0.263$,
$+0.311$), while on HT-reference data it reverses --- the reference-alignment mechanism of
Appendix~\ref{app:refdep}. And \est{IPS} over-selects the easiest-to-evaluate candidate
($\sim\!1.8\times$ its true-best rate on the easy slate, up to $2.8\times$ on the
competitive one, \est{DM} the same at design-dependent strength;
Table~\ref{tab:rq4-decoy}) --- a pathology of degree.
(\est{Perturbation-DR} is excluded; Appendix~\ref{sec:smoothing}.)

\subsection{An external check on a non-simulated reference}
\label{sec:twins-main}

One check stands apart from the robustness suite because its reference is not a surface
anyone modelled. On the \dataset{Twins} cohort --- $11{,}400$ same-sex twin pairs with
both siblings' outcomes recorded, taking the pair as the unit --- the RQ1 ordering
replicates at a \emph{larger} margin ($5.0\times$ against $2.0\times$), the RQ2
alignment mechanism replicates, and the RQ3 cross-fitting sign replicates. The calibrated
cut points do not transfer. That is exactly the split Section~\ref{sec:cannot} predicts:
mechanisms travel, magnitudes do not. Appendix~\ref{app:twins} gives the design, the
matched-pair caveat, and the full account of what fails to transfer.

% ============================================================
\section{A Practitioner's Guide}
\label{sec:guide}

\begin{enumerate}[leftmargin=*,itemsep=5pt]
\item \textbf{Ask how far your target is from your logger before trusting any
  importance-weighted estimate.} A log whose policy broadly agrees with the rule you want
  to deploy supports weighting even under sharp logging; one that ignores it does not
  (RQ2). ESS fraction and support deficiency work at their demonstrated scope --- ranking
  \emph{logging situations}, not candidates within one log (median $\rho=-0.11$) --- and
  our cut points ($0.36$, $0.02$) describe this benchmark: read a low ESS as a reason to
  seek a better-aligned log or a model-based estimator, not a go/no-go test. Use the
  screen ex ante: to choose which log to collect or buy, and to judge whether an
  evaluation \emph{program} on a given log--target pair is viable at all. It does not
  rank candidates within the log you already hold. \textbf{Fit
  and check a flexible propensity model first}: a poor one leaves the screen uninformative
  or inverted without rescuing \est{IPS} accuracy (Section~\ref{sec:propensity}).

\item \textbf{Prefer DR under model uncertainty; DM is competitive when adequacy is
  demonstrable.} \est{DM} is best or tied-best on four of five datasets under in-sample
  nuisances but only three of five with out-of-fold ones, where it is the worst of six on
  \dataset{Jobs} and \dataset{Lenta} (Table~\ref{tab:rq1-oof}); it leads the exact-value
  cross-dataset mean under both and neither HT-reference mean. The case for \est{DR} is
  \emph{asymmetric risk}: under a weak outcome model \est{DM} can degrade to worst-of-four
  while \est{DR} degrades gracefully (Section~\ref{sec:omquality}) --- though that rests
  on one DGP family of three, although the crossing itself survives seed resampling
  (\est{DM}$-$\est{IPS} gap $[+0.022,\,+0.063]$ at the stump rung). Judge adequacy by
  held-out error \emph{per arm} --- and trust the intervals less than the points: the
  reported CIs are conditional on the fitted nuisance and undercover for \est{DM}
  ($0.89$ against nominal $0.95$), with weighting coverage degrading in exactly the
  flagged low-overlap regime (Appendix~\ref{app:coverage}).

\item \textbf{Against the optimizer's curse, decide by data provenance --- do not try to
  detect the bias.} If the policy was learned from the outcomes you evaluate on, use
  honest policy-level splitting; if trained on an independent sample, plain DR needs no
  policy-level split for reuse bias. Nuisance-only cross-fitting \emph{worsened} it (RQ3).

\item \textbf{When \emph{selecting} among candidates, rank them all on one shared log.}
  Scoring each candidate on a log aligned with itself compares policy--logger pairs, not
  policies (RQ4). One robust pathology: \est{IPS} over-selects the easiest-to-evaluate
  candidate --- up to $2.8\times$ the decoy's true-best rate, \est{DM} matching it under
  the common-independent logger (Table~\ref{tab:rq4-decoy}).

\item \textbf{Always show per-dataset and per-regime results alongside any aggregate.}
  Pooling exact-value with noisy-reference results hides mechanism.
\end{enumerate}

\section{Conclusion}

Budget-constrained allocation is where naive offline evaluation misleads most. Weak
overlap is governed by logger--target \emph{action} alignment, not by sharpness, and is
rankable in advance \emph{across} logging environments --- not within one log, and only
with a good propensity model. Under policy--evaluation reuse, cross-fitting the nuisance
alone makes the optimism \emph{worse}; honest policy-level splitting avoids the reuse by
targeting the learning procedure's value, not by de-biasing the full-sample policy's. The mechanisms --- though not the calibrated cut points ---
replicate against a non-simulated paired reference we did not construct
(Appendix~\ref{app:twins}).
Doubly-robust estimation is the most stable across our stresses,
Section~\ref{sec:guide} is the operational form, and every number regenerates from the
released repository's documented \texttt{make} targets
(\url{https://github.com/binshuangli/allocation-ope-bench}).

\section*{Broader Impact Statement}
\looseness=-1 This is protective methodology: offline estimates of allocation policies
can be confidently wrong, and we supply diagnostics for when not to trust them and when
those diagnostics fail. The main risk is reading the screen as a certificate; all data
is public. \textbf{Use of generative AI.}\enspace Generative AI assisted with writing
and code; the authors take full responsibility.

\bibliography{refs}
\bibliographystyle{tmlr}

\appendix

% ============================================================
\section{Claims and supporting evidence}
\label{app:claims}

\begin{table}[H]
\centering
\small
\begin{tabular}{p{0.62\linewidth}l}
\toprule
Claim & Evidence \\
\midrule
Overlap is the strongest driver of weighting error \emph{among the logging-design axes}
(propensity error is larger) &
  Fig.~\ref{fig:rq2-alignment}, \S\ref{sec:rq2} \\
The model-based advantage over IPS is the robust effect (paired bootstrap, CI
excludes 0); the \est{DM}-over-\est{DR} margin is tiny, partly baseline-driven,
and not distinguishable from zero under nuisance-refit resampling &
  \S\ref{sec:rq1real} \\
ESS fraction and support deficiency rank realized IPS error, consistently
within every dataset and under leave-one-dataset-out --- but \emph{across} logging
environments, not within one fixed log, where the median within-log correlation is
$-0.11$ over $450$ logs and no better than chance under a candidate-independent
logger &
  Fig.~\ref{fig:rq2-trust}, \S\ref{sec:rq2}, App.~\ref{app:cluster} \\
The risk ranking transfers to held-out DGP families (ESS ROC-AUC $0.83$ at a $10\%$
error target, and $0.91$ at the $2\%$ target matched to the larger suite's error scale) &
  \S\ref{sec:rq2}, \S\ref{sec:robustness} \\
The reuse-bias sign reduces to an exact finite-sample identity whose expectation,
under the benchmark's known propensities, is $-\mathrm{Cov}(w\,\mathbb{1}[a{=}\pi_e],
\hat\mu_{\text{in}})$; whether the covariance is material is the tested content &
  \S\ref{sec:rq3} \\
Model-based estimators inherit outcome-model error; \est{DM} can become worst
under severe misspecification (seed-resampled: gap $[+0.022,\,+0.063]$),
\est{DR} degrades gracefully &
  Table~\ref{tab:misspec}, \S\ref{sec:omquality}, App.~\ref{app:robustness} \\
No transferable model-adequacy screen follows from factual held-out RMSE &
  \S\ref{sec:omquality} \\
Nuisance-only cross-fitting does not remove reuse bias; honest policy-level
splitting cuts it $58$--$92\%$ (8 regimes) &
  Fig.~\ref{fig:rq3}, Table~\ref{tab:rq3-debias}, \S\ref{sec:rq3} \\
Material optimizer bias appears only on the continuous known-effect datasets in this benchmark &
  Table~\ref{tab:rq3-debias}, \S\ref{sec:rq3} \\
In correct-selection accuracy the \est{DM}--\est{IPS} gap stays positive across logging
designs ($+0.056$ to $+0.105$ on the three-candidate slate) and shrinks on the
competitive slate; normalized regret agrees on the three-candidate slate but
\emph{reverses} on the seven-policy slate, driven by the HT-reference datasets.
Selection conclusions therefore depend more on the candidate slate and the evaluation
metric than on the logging design &
  Table~\ref{tab:rq4-logger}, \S\ref{sec:rq4} \\
IPS over-selects the easiest-to-evaluate candidate under all three logging designs
($\sim\!1.8\times$ its true-best rate on the three-candidate slate, more on the
competitive one); DM shows the same tendency, generally more weakly under the aligned
designs but similarly under the common-independent logger &
  \S\ref{sec:rq4}, Table~\ref{tab:rq4-decoy} \\
The estimator ordering survives misaligned logging \emph{qualitatively}; against a
\emph{disjoint} RCT reference it shifts --- the model-based advantage strengthens and the
weighting advantage disappears --- so RCT-reference comparisons cannot resolve it &
  \S\ref{sec:robustness}, App.~\ref{app:refdep} \\
Findings replicate on a second, independent real covariate set &
  \S\ref{sec:robustness} \\
The mechanisms (RQ1 ordering, alignment-governed overlap, screen association, reuse-bias
sign) replicate against a reference read off recorded paired outcomes; the failure
criterion and error levels do not transfer &
  App.~\ref{app:twins} \\
\bottomrule
\end{tabular}
\caption{Each headline claim and the section, table, or figure that supports it.}
\label{tab:claims}
\end{table}

% ============================================================
\section{Diagnostic mechanics, estimator addenda, and reproduction targets}
\label{app:mechanics}

\paragraph{Score alignment is not action alignment: the correct sharpening limit.}
Let $\pi_e$ be top-$k$ under score $s$ with standardized cutoff $\tilde q>0$, and let the
score-aligned logger be
$\pi_b(1\mid x)=\mathrm{clip}\big(\sigma(\tilde s(x)/\tau);\,\varepsilon,\,1-\varepsilon\big)$,
$\varepsilon=0.02$. Writing $q(X)=\pi_b(a_{\pi_e}(X)\mid X)$ for the logging probability
of the \emph{target's} action, the weight at the logged action satisfies
$\mathbb E[W\mid X]=1$ and $\mathbb E[W^2\mid X]=1/q(X)$, so
$\mathrm{ESS}/n \to 1/\mathbb E[1/q(X)]$. As $\tau\to0$, $q(X)\to1-\varepsilon$ on the
aligned mass $c=k+F_{\tilde s}(0)$ but $\to\varepsilon$ on the band
$0<\tilde s<\tilde q$ between the score mean and the budget cutoff, where the logger
treats and the target does not. Hence
\[
\frac{\mathrm{ESS}}{n} \;\longrightarrow\;
\Big[\frac{c}{1-\varepsilon}+\frac{1-c}{\varepsilon}\Big]^{-1}
\;\approx\; \frac{\varepsilon}{1-c},
\]
which is small: with $k=0.1$ (so $c\approx0.6$) the limit is $\approx0.05$. The
floor-probability draws occur with probability $O(\varepsilon)$ but carry weight
$1/\varepsilon$, an $O(1/\varepsilon)$ second-moment contribution that cannot be
neglected. The case $\tilde q>0$ covers budgets smaller than the mass above the score
mean; in general (continuous scores) the mismatch band lies between the mean and the
cutoff on whichever side the cutoff falls,
$c=1-\lvert F_{\tilde s}(\tilde q)-F_{\tilde s}(0)\rvert$ --- reducing to
$k+F_{\tilde s}(0)$ for $\tilde q>0$ --- so the same limit covers the benchmark's full
budget grid. \emph{Sharpening a score-aligned logger therefore eventually collapses
overlap}: the flatness of Figure~\ref{fig:rq2-alignment} is a finite-range plateau, and
the collapse begins just below the tested range. Extending the temperature grid downward
confirms this on the benchmark's own data, and an \emph{action-aligned} logger --- the
same logistic centred at the target's top-$k$ cutoff, for which the mismatch band is
empty --- behaves oppositely, sharpening toward full support (limit $1-\varepsilon$):

\begin{center}\small
\begin{tabular}{lrrrrrr}
\toprule
median ESS at $k{=}0.1$ (not budget-pooled) & $\tau{=}5$ & $2$ & $0.5$ & $0.25$ & $0.1$ & $0.05$ \\
\midrule
score-aligned  & 0.512 & 0.519 & 0.398 & 0.171 & 0.076 & 0.061 \\
action-aligned & 0.560 & 0.632 & 0.836 & 0.907 & 0.946 & 0.960 \\
\midrule
median $|$IPS rel.\ bias$|$ (pooled $k$) & & & & & & \\
score-aligned  & 0.026 & 0.026 & 0.028 & 0.029 & 0.049 & 0.065 \\
action-aligned & 0.025 & 0.028 & 0.022 & 0.020 & 0.019 & 0.017 \\
\bottomrule
\end{tabular}
\end{center}

\noindent(\texttt{make repro-sharpening-limit}.) Temperature is therefore not a valid
overlap parameter in either direction: what governs support is the logger's probability
of the \emph{target's actions}, and the main sweep's temperature range happens to sit on
the score-aligned plateau --- one more reason the alignment axis, not sharpness, is the
right design variable for benchmarks of this estimand.

\textbf{What the three diagnostics measure.} Write
$w_i=\pi_e(a_i\mid x_i)/\pi_b(a_i\mid x_i)$ for the importance weight. The \emph{ESS
fraction} is Kish's effective sample size normalized by the logged sample size,
$\big(\sum_i w_i\big)^2\big/\big(n\sum_i w_i^2\big)$. \emph{Support deficiency} is the
fraction of evaluation units whose target-selected action was taken by the logger with
probability below $\kappa\!=\!0.05$ --- the share of the target policy's decisions that
the log barely covers. \emph{Max weight} is $\max_i w_i$, the largest realized importance
weight --- bounded by $50$ here by the propensity floor. The quantity all
three are asked to rank is IPS $|$relative bias$|$,
$|\Vhat-V_{\mathrm{ref}}|/|V_{\mathrm{ref}}|$, taken per cell as the \emph{median over its
ten seeds} --- a signed-error magnitude, as opposed to the relative RMSE of Section~\ref{sec:rq1}, which additionally
averages over seed variance. The correlations above are computed over all $540$
cells of the alignment sweep, \emph{not} over the main accuracy sweep.

\textbf{ESS fraction is not a pure dispersion measure here, and it matters why.} Under a
\emph{deterministic} target $w_i\in\{0,1/\pi_b(a_i\mid x_i)\}$, so units whose logged
action differs from the one the target selects contribute zeros that remain in the
normalizer. Writing $m$ for the number of \emph{matched} units, the statistic factors
exactly as
$(m/n)\times\big[(\sum_{\text{matched}} w)^2/(m\sum_{\text{matched}} w^2)\big]$ --- an
action-\emph{coverage} term times a dispersion term on the matched set. It therefore
moves with the budget and the logger's marginal treat rate, not with weight dispersion
alone. Two consequences we flag rather than paper over: the value $1$ is unattainable by
construction, and the numeric cut point used in Section~\ref{sec:guide} is not portable
across budgets, loggers or sample sizes even within this benchmark. What the diagnostic
buys is a \emph{ranking} of cells within a comparable design, which is what
Table~\ref{tab:rq2-conditioning} conditions on.

\textbf{Support deficiency rests on two nested constants} --- $\kappa$, and then the
screen's threshold on the resulting fraction --- so it deserves a sensitivity
note. Recording it at $\kappa\in\{0.01,0.05,0.10\}$, the association is stable between
$0.05$ and $0.10$ ($\rho=+0.42$ and $+0.43$; all diagnostics are
seed-averaged per cell, as everywhere else in this section), while $\kappa\!=\!0.01$ is degenerate
here: the $0.02$ propensity floor (Section~\ref{sec:design}) means no unit can fall below
it --- the same floor that gives weight clipping no purchase
(Section~\ref{sec:rq1real}). The floor also makes support deficiency a
floor-\emph{dependent} quantity (mass in $[0.02,\kappa)$) where ESS fraction is
floor-robust, which is one reason we expect the latter to be the half of the screen that
transfers.

\textbf{Does weight clipping rescue IPS?} Truncating importance weights is the standard
practitioner response to exactly this pathology, so we added clipped IPS
($\min(w_i,M)$ for $M\in\{5,10,50\}$) on the exact-value datasets
(\dataset{synthetic}, \dataset{IHDP}, and one ACIC nonlinear surface, on the reduced
three-budget grid of the robustness runs --- absolute levels therefore differ from
Table~\ref{tab:rq1-accuracy}, as in Appendix~\ref{app:refdep}). Because our own
Section~\ref{sec:rq1real} argument is that medians hide the tail where weighting fails, we
score clipping on tail-sensitive summaries, where it would show up if it worked. It does
not: the mean paired difference against raw \est{IPS} is $+0.0000$ at $M{=}50$,
$-0.0000$ at $M{=}10$, and $+0.0005$ at $M{=}5$ (slightly \emph{worse} --- clipping bias
appearing before any variance benefit); the 90th and 99th percentiles of per-cell error
are unchanged at $M{\geq}10$ and the 99th \emph{worsens} at $M{=}5$ ($0.078 \to 0.085$);
the single worst cell is untouched; and only $11\%$ of cells are affected at all even at
$M{=}5$. The reason is structural. Our logging propensities are floored at $0.02$
(Section~\ref{sec:design}), so weights are bounded by $50$ by construction, and the
median per-cell maximum weight is only $5.0$ at $\tau{=}0.5$ on the main accuracy
sweep, against a median ESS fraction of $0.56$ in the same cells (the worst single cell reaches $0.084$, but that is
a minimum, not the typical case). The failure mode is \emph{not} a handful of enormous
weights that truncation would catch; it is the ordinary dispersion of \emph{many}
moderately inflated weights, which is bounded below by the maximum weight --- at
$W\!=\!5.0$ no reweighting scheme that only touches the top of the distribution can lift
the dispersion term materially above where it already sits. The natural objection --- that this
experiment, like the main sweep, uses a self-aligned logger where weights stay modest ---
is now closed directly: rerunning clipping under all three logger regimes on the
exact-value datasets (\texttt{make repro-tail-control}), including the independent logger
where maximum weights reach the $50$ ceiling, the paired median change against raw
\est{IPS} is $0.0000$ at every $M$, with $M{=}10$ helping in at most $25\%$ of cells.
Clipping is ineffective even where the tail exists \emph{under this floor} --- and that
qualifier carries the result. Sweeping the floor itself
(\texttt{make repro-floor-sensitivity}; $\varepsilon$ to $0.0002$, ceilings to $5{,}000$,
crossed with $\tau$ to $0.05$), clipping is a no-op at $\varepsilon{=}0.02$ in $100\%$ of
cells at $M{=}50$ --- the floor has already clipped --- but useful once weights exceed
that ceiling (median \est{IPS} $0.111$ against $0.090$ at $\varepsilon{=}0.005$), while
\est{DR} stays flat ($0.038\to0.037$). \emph{``Clipping does not help'' is a property of
the bounded-weight regime we chose, not of allocation OPE.} Large maximum weights do
\emph{co-move} with realized error --- max weight is a valid marker, positive in all five
per-dataset strata of the alignment sweep ($+0.29$ to $+0.72$) --- but the marker is a symptom of the same
dispersion that collapses ESS, not a separable cause that clipping can remove. Shrinkage-DR \citep{su2020doubly}, which also targets the weight tail, is now
evaluated in the same rerun: with its $\lambda$ tuned on the logged sample by the same
estimated-MSE style as the other hybrids, it edges plain \est{DR} by a median $0.0003$--$0.0008$
relative RMSE ($67$--$74\%$ of cells, largest under the independent logger) --- a real
but small improvement, consistent with the expectation that tail control buys little when
the pathology is ESS collapse. Practitioners should not expect weight control to substitute for overlap \emph{when
weights are already bounded}; where they are not, clipping is worth using, and the floor
sweep above locates that boundary.

\textbf{Why the two tuned hybrids duplicate their untuned parents.} In
Table~\ref{tab:rq1-accuracy} \est{Switch-DR} sits within $0.0003$ of \est{DR} and
\est{mIPS} within $0.001$ of \est{IPS} on every dataset, which invites the suspicion that
the tuning is doing nothing. Recording the \emph{selected} hyper-parameter in all
$2{,}700$ configurations (\texttt{make repro-tuner-selection}) shows that is close to
right, and identifies the mechanism as a property of the selection rules rather than a
coincidence. \est{Switch-DR} never selects $\lambda_{\mathrm{sw}}=\infty$, but the $\lambda_{\mathrm{sw}}$ it does select
exceeds the cell's \emph{largest} importance weight in $94.9\%$ of configurations: no unit
is switched, so the estimator is numerically identical to \est{DR} there --- close to the
$96.6\%$ of cells in which the two estimates agree bit-for-bit. The propensity
floor is again the reason. With weights bounded by $50$ and a median per-cell maximum of
$2.1$ (at $\tau{=}5$) to $5.0$ (at $\tau{=}0.5$), the grid's \emph{smallest} candidate
$\lambda_{\mathrm{sw}}\!=\!5$ --- selected $83\%$ of the time --- already sits above most of the weight
distribution. \est{mIPS} likewise selects $\alpha=0$, which reduces it to \est{IPS}
exactly, in $75.0\%$ of configurations. The residual $25.0\%$ is not negligible per cell
(median $|\Delta\Vhat|/|V|$ of $0.006$, 90th percentile $0.080$), so \est{mIPS} is not a
relabelling of \est{IPS} configuration by configuration; those cells are simply too few to
move any dataset's median. The general point is that both estimated-MSE proxies collapse
to the untuned estimator under precisely the conditions where tuning would have to earn
its keep: neither can buy variance reduction when the weight tail is bounded by
construction --- the same reason clipping fails below.

\subsection{Reproduction targets}
\label{app:repro-targets}
\begin{figure}[t]
  \centering
  \includegraphics[width=0.85\linewidth]{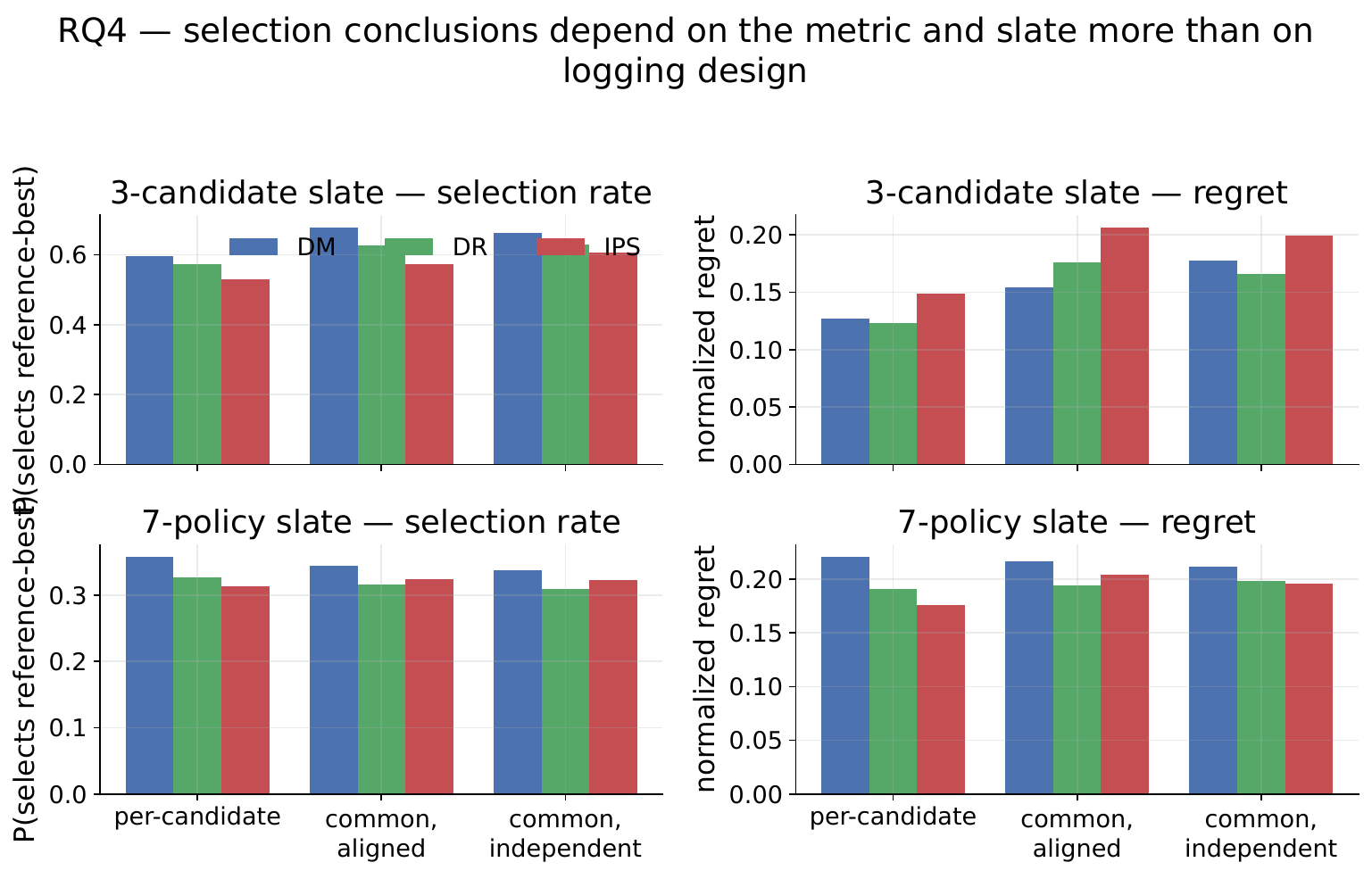}
  \caption{RQ4 --- selection quality under three logging designs, on both slates
  (top: three candidates; bottom: seven). Left: rate of selecting the reference-best
  policy (higher better). Right: mean normalized regret (lower better). Moving left to
  right --- from a per-candidate log, to one shared log aligned with a single candidate,
  to a shared candidate-independent log --- the bars move little, which is the point: the
  family ordering in correct-selection accuracy is largely insensitive to the logging
  design, while it weakens sharply
  between the top and bottom rows as the slate becomes competitive. These are
  \emph{pooled} rates; Section~\ref{sec:rq4} shows
  the pooled null averages a real model-based advantage on exact-value data with a
  reversal on HT-reference data, so the figure should be read together with that split.
  On the seven-policy slate normalized regret (right column) \emph{reverses} the pooled
  \est{DM}--\est{IPS} ordering, placing \est{IPS} lower; both exact-value datasets still
  favour \est{DM}.}
  \label{fig:rq4}
\end{figure}

\begin{itemize}[leftmargin=*,itemsep=0pt]
\item \texttt{make repro-full} --- the $2{,}700$-configuration accuracy run
  ($18{,}900$ estimator-level results $=16{,}200$ evaluated outputs from the six
  common-estimand estimators plus $2{,}700$ perturbation-DR rows excluded from the
  common-estimand tables; their matched-reference evaluation is
  \texttt{make repro-perturbation-matched}; RQ1--RQ2, RQ4);
\item \texttt{make repro-optbias} --- the optimization-bias study (RQ3);
\item \nopagebreak \texttt{make repro-acic} and \texttt{make repro-acic-hillstrom} --- the two
  known-effect \emph{temperature-design} hardening sweeps ($6$ DGPs $\times$ $3$
  candidates $\times$ $4$ budgets $\times$ $3$ temperatures $\times$ $10$ seeds
  $=2{,}160$ configurations each). Their \emph{alignment-axis} counterparts,
  \texttt{make repro-logger-alignment-acic} and
  \texttt{make repro-logger-alignment-acic-hillstrom}, add the three logger regimes but
  need two learned candidates rather than three (each serves as the other's misaligned
  logger), so they come to $4{,}320$ configurations each rather than $6{,}480$;
\item \texttt{make repro-twins} --- the non-simulated paired-reference validation of
  the RQ1--RQ3 mechanisms (Appendix~\ref{app:twins});
\item \texttt{make reference-check} --- numerical agreement of \est{IPS}, \est{SNIPS},
  \est{DM} and \est{DR} with Open Bandit Pipeline on identical inputs;
\item \texttt{make repro-misspec} --- the outcome-model degradation study;
\item \texttt{make repro-selection} --- the seven-policy selection slate;
\item \texttt{make repro-commonlog} and \texttt{make repro-commonlog-3cand} --- the
  common-logger RQ4 experiment on the seven- and three-candidate slates
  (Table~\ref{tab:rq4-logger}, Figure~\ref{fig:rq4}), which is the primary selection
  evidence;
\item \texttt{make repro-full-cutoff} --- the whole accuracy sweep and alignment sweep
  re-run with the logistic centred at each budget's top-$k$ cutoff (the cutoff-centred
  design);
\item \texttt{make repro-full-oof} --- the whole accuracy sweep re-run with out-of-fold
  outcome nuisances (the last two columns of Table~\ref{tab:rq1-accuracy});
\item \texttt{make repro-nuisance-crossfit} --- the out-of-fold-nuisance robustness
  check behind the RQ1 family claim (Section~\ref{sec:rq1real});
\item \texttt{make repro-tuner-selection} --- the selected \est{Switch-DR} $\lambda_{\mathrm{sw}}$ and
  \est{mIPS} $\alpha$ over the main grid (Section~\ref{sec:rq1real});
\item \texttt{make repro-tail-control} --- clipping and shrinkage-DR under all three
  logger regimes;
\item \texttt{make repro-sharpening-limit} --- score- vs action-aligned logging down to
  $\tau{=}0.05$ (the action-aligned overlap limit);
\item \texttt{make repro-floor-sensitivity} --- the propensity-floor sweep
  ($\varepsilon$ down to $0.0002$) behind the scope of the tail-control finding;
\item \texttt{make repro-optbias-decoupled} --- the policy/nuisance independence
  ablation behind the RQ3 sign;
\item \texttt{make repro-refit-intervals} --- refit-aware intervals (Appendix~\ref{app:coverage});
\item \texttt{make repro-logger-alignment} --- the alignment-axis RQ2 sweep
  (Section~\ref{sec:rq2});
\item \texttt{make repro-logger-alignment-acic} and
  \texttt{make repro-logger-alignment-acic-hillstrom} --- the alignment-axis held-out
  suites that carry the screen AUCs;
\item \texttt{make repro-rct-disjoint} --- the shared-vs-disjoint reference split on all
  three RCT datasets (Appendix~\ref{app:refdep});
\item \texttt{make analyze} / \texttt{make analyze-acic} --- every figure and table
  in this paper, regenerated from the parquets.
\end{itemize}

\section{Robustness checks and the outcome-model ladder: design and per-DGP detail}
\label{app:moved}

\subsection{Further robustness checks}
\label{sec:robustness}

Three checks probe whether the RQ1--RQ2 conclusions depend on the benchmark's
construction; none suggests they do. Full tables are in Appendix~\ref{app:robustness}.

\textbf{Known-effect hardening on real covariates.} The exact-value evidence so far
rests on one synthetic generator and one semi-synthetic dataset. We add six known-effect
DGPs built on the \emph{real} \dataset{IHDP} covariates, and repeat the whole sweep on a
second, independent covariate source (a $10{,}000$-row \dataset{Hillstrom} subsample) ---
$2{,}160$ further configurations, zero anomalies. The RQ1 ordering replicates in full on
both: \est{DM} most accurate, the DR family next, \est{IPS}/\est{mIPS} worst \emph{in
every setting} ($2.1\times$ \est{DM} on the \dataset{IHDP}-covariate sweep and $3.2\times$ on the \dataset{Hillstrom}-covariate sweep). The RQ2 diagnostics
replicate independently too: on the \emph{alignment-axis} hardening suites --- separate reruns of the same six DGPs
($4{,}320$ configurations each, in addition to the two $2{,}160$-configuration
temperature-design sweeps; these reruns carry the held-out screen AUCs of $0.83$ and
$0.91$), $\rho(\text{ESS})=-0.45$ and $-0.58$, versus $-0.41$ on the primary sweep; the
temperature-design suites of Appendix~\ref{app:robustness} give $-0.37$ and $-0.52$. The separate alignment-axis suites provide the held-out validation results and AUCs
reported in Section~\ref{sec:rq2}.

\textbf{A non-simulated reference.} Both checks above still score against simulated
response surfaces. The \dataset{Twins} cohort supplies an evaluation
reference read directly off recorded paired outcomes instead
(Appendix~\ref{app:twins}). The RQ1 ordering holds with a \emph{larger} model-based
margin ($5.0\times$ against $2.0\times$), the alignment mechanism and the screen's
association replicate, and the RQ3 signs replicate; the $10\%$ failure criterion and the
absolute error levels do not transfer, which is the split
Section~\ref{sec:cannot} predicts.

\textbf{Misaligned logging.} The main sweep logs under a score-aligned stochastic logger.
Crossing logger and target (log under the \est{T-learner}, evaluate the \est{S-learner},
and vice versa) raises every estimator's error modestly --- model-based $+12.3\%$, IPS
family $+15.4\%$ --- and preserves the family ordering. This is a check on the
\emph{ordering}, not on the size of the gap: crossing two correlated meta-learners is a
weaker perturbation than the candidate-independent logger of Section~\ref{sec:rq4}, under
which we did not re-score accuracy, so the RQ1--RQ2 magnitudes remain
self-aligned-logger quantities (Appendix~\ref{app:align}).

\paragraph{The outcome-model ladder.}\label{app:omquality} Design and per-DGP
detail are in Appendix~\ref{app:misspec}; the finding is stated in
Section~\ref{sec:omquality}.

\section{RQ4: reference-type split and decoy detail}
\label{app:rq4-extra}

\textbf{Splitting by reference type shows the pooled null hides two opposing effects.}
Our own rule --- never pool exact-value and HT-reference results in a single claim ---
applies here too. Under the common independent logger the split is reported in the last
block of Table~\ref{tab:rq4-logger}. On the \emph{exact-value} datasets --- where ``correct'' means recovering the genuinely
value-maximizing candidate --- the model-based advantage is large and intact even under a
shared, candidate-independent log ($+0.263$ and $+0.311$). On the \emph{HT-reference}
datasets it reverses: \est{IPS} matches or beats \est{DM}. We do not read the reversal as
evidence that weighting selects better. The HT reference is itself an inverse-propensity
construction computed on the same randomized split, so \est{IPS} and the reference share
an estimator family and much of their noise; agreement between them is partly structural
rather than evidence of accuracy. The pooled RQ4 null is therefore the average of a real
model-based advantage where truth is exact and what we suspect is a reference-alignment
artifact where it is not. The defensible statement is narrower than either half: on a
competitive slate scored against a shared independent log, we can separate the families
only where exact ground truth is available, and we cannot rule out that the apparent
parity on RCT data is an artifact of the reference construction.

One dataset drives much of the instability. \dataset{Jobs} has a \emph{negative}
$\est{DM}-\est{IPS}$ gap in \emph{all six} designs (per-dataset gaps are tabulated in
Appendix~\ref{app:rq4-percand}; $-0.113$ to $-0.367$). That is coherent with the accuracy results:
\dataset{Jobs} is the one dataset where \est{IPS} is nominally best in
Table~\ref{tab:rq1-accuracy}, and Section~\ref{sec:rq1} attributes that to its small
retained sample rather than to any \est{IPS} advantage.

\section{The cutoff-centred logger}
\label{app:cutoff}

\textbf{The cutoff-centred design confirms the analysis.} Our logistic is centred at the
score \emph{mean}, so the self-aligned logger matches the candidate's score ranking, not
its action boundary. Re-running the benchmark and this sweep centred at each budget's
top-$k$ cutoff (\texttt{make repro-full-cutoff}) confirms both predictions: sharpening an
action-aligned logger \emph{improves} overlap (median ESS $0.545 \to 0.766$ as $\tau$
falls, against $0.522 \to 0.562$ under mean-centring), and the gradient sharpens to
$0.0\%$ / $1.7\%$ / $26.7\%$ failure at $\tau{=}0.5$ (screen AUC $0.954$). That pooled
AUC is \emph{higher} than the primary sweep's while the pooled $\rho$ of
Section~\ref{sec:rq2} is lower, and the reconciliation is composition: failures here
concentrate almost entirely in the independent regime ($16$ of $17$ failing cells), so a
pooled AUC partly measures regime separation --- within the independent regime it is
$0.869$ --- which is why the comparison in Section~\ref{sec:rq2} leads with $\rho$. The
drop in pooled $\rho$ ($-0.41 \to -0.32$) is likewise what removing the self-aligned
stratum's failures implies: fewer positives to rank, not a weaker screen. The residual
self-aligned failures were an artifact of mean-centring; the estimator ordering is
unchanged and every error falls (\est{IPS} $0.065\to0.052$). We keep the mean-centred
sweep primary for a substantive reason: it spans a \emph{wider overlap range}, which is
what a study of a risk-\emph{ranking} diagnostic needs. Under the cutoff-centred logger the
self-aligned cells essentially never fail, leaving the screen little to discriminate --- a
better logging design makes a worse testbed for a fragility diagnostic.

\section{Why the screen inverts under a constant propensity model}
\label{app:propensity-mech}

Section~\ref{sec:propensity} reports that under the marginal $\hat\pi_b$ the fragility
screen is not merely uninformative but \emph{inverted} (AUC $0.05$). The stratum
numbers show why this is systematic rather than noise. Under that model the
self-aligned $\tau{=}0.5$ cells --- the ones whose true propensities are most extreme,
and therefore where a constant estimate is most wrong --- carry simultaneously the
\emph{highest} estimated ESS fraction ($0.663$) and the \emph{highest} \est{IPS} error
($0.467$). A constant $\hat\pi_b$ contributes no dispersion, so the estimated ESS
reduces to a coverage measure, and coverage is maximized exactly where the sharp,
score-aligned logger agrees most often with the target. The diagnostic therefore ranks
the worst cells as the safest. This is the concrete sense in which fitting a flexible
propensity model is a prerequisite for the screen rather than a refinement of it.

\section{RQ2 conditioning ladder: table and detail}
\label{app:rq2-cond-full}

\begin{table}[H]\centering
\caption{The full $\delta$ grid behind the held-out screen validation: ROC-AUC of
$\widehat{\mathrm{ESS}}$ for predicting median \est{IPS} $|$relative bias$| > \delta$ on
the two alignment-axis hardening suites ($432$ cells each), with the share of cells above
each $\delta$. Dashes mark degenerate strata (no cells, or essentially all cells, above
$\delta$). The pre-specified $10\%$ target is scoreable only on the IHDP-covariate suite;
on the Hillstrom-covariate suite no cell reaches it, which is why that suite is scored at
the $2\%$ target matched to its error scale --- chosen for scale, not selected for AUC.}
\label{tab:delta-grid}
\small
\begin{tabular}{lccccc}
\toprule
 & $\delta{=}1\%$ & $2\%$ & $5\%$ & $10\%$ & $20\%$ \\
\midrule
IHDP-covariate suite: AUC & -- & 0.68 & 0.72 & \textbf{0.83} & -- \\
\quad share of cells above $\delta$ & $100\%$ & $97\%$ & $59\%$ & $17\%$ & $0\%$ \\
\addlinespace
Hillstrom-covariate suite: AUC & 0.77 & \textbf{0.91} & -- & -- & -- \\
\quad share of cells above $\delta$ & $69\%$ & $15\%$ & $0\%$ & $0\%$ & $0\%$ \\
\bottomrule
\end{tabular}
\end{table}

\begin{table}[H]\centering
\caption{RQ2 diagnostic association under progressively stricter conditioning
(\est{IPS}), on the logger-alignment sweep. All quantities are computed over \emph{cells}
--- the seed-aggregated unit of Section~\ref{sec:design} --- so repeated seeds do not
inflate the sample; AUC is the screen's discrimination for the $|$relative
bias$|>10\%$ target and ``unsafe'' the share of cells exceeding it. The ranking survives
conditioning on dataset (negative in $5/5$) and on budget (AUC $0.68$--$0.95$ in every
stratum), and survives conditioning on the logger regime where failures actually occur
(independent logger, $\rho=-0.56$ with $11.7\%$ unsafe); it does \emph{not} survive
conditioning on regime and $\tau$ jointly (median $-0.01$, negative in $6/9$). We read the
last row as expected rather than damaging: regime $\times$ $\tau$ is the mechanism that
generates overlap variation here, so fixing both removes the variation the diagnostic
exists to rank. The defensible claim is between-situation ranking --- which logging
situation am I in --- not within-situation discrimination.}
\label{tab:rq2-conditioning}
\small
\begin{tabular}{lrrrr}
\toprule
conditioning & cells & $\rho(\text{ESS})$ & AUC(ESS) & unsafe \\
\midrule
pooled (all cells)                    & 540 & $-0.41$ & $0.85$ & $6.3\%$ \\
\midrule
\multicolumn{5}{l}{\emph{within logger regime (3 strata of 180 cells):}} \\
\quad independent (logger ignores target) & 180 & $-0.56$ & $0.75$ & $11.7\%$ \\
\quad misaligned (other candidate)        & 180 & $-0.17$ & $0.74$ & $4.4\%$ \\
\quad self-aligned (score-aligned; same candidate score)    & 180 & $-0.27$ & $0.97$ & $2.8\%$ \\
\midrule
\multicolumn{5}{l}{\emph{within dataset (5 strata of 108 cells):}} \\
\quad range over datasets             & 108 & $-0.35$ to $-0.71$ & --- & --- \\
\quad leave-one-dataset-out, pooled   & 432 & $-0.37$ to $-0.45$ & --- & --- \\
\midrule
\multicolumn{5}{l}{\emph{within budget $k$ (6 strata of 90 cells):}} \\
\quad range over budgets              & 90 & $-0.22$ to $-0.56$ & $0.68$--$0.95$ & $2.2$--$15.6\%$ \\
\midrule
\multicolumn{5}{l}{\emph{within regime \textbf{and} $\tau$ (9 strata of 60 cells):}} \\
\quad median over the 9 strata        & 60 & $-0.01$ (6/9 neg.) & --- & --- \\
\bottomrule
\end{tabular}
\end{table}

\textbf{How much of this is our own design?} The variation the diagnostic ranks is
\emph{induced} by the two axes we control --- the logger regime and the temperature
$\tau$ --- so the association could in principle be recovering nothing more than ``which
experimental cell am I in'', which is useless to a practitioner holding one logged
dataset and no dials. Table~\ref{tab:rq2-conditioning} works through the conditioning
ladder, and the answer is partly yes.

Conditioning on \emph{dataset} leaves the association intact ($-0.35$ to $-0.71$,
negative in $5/5$), as does conditioning on \emph{budget} ($\rho=-0.22$ to $-0.56$, AUC
$0.68$--$0.95$ across strata), so the ranking is not an artifact of pooling across
datasets of very different size or across budget difficulty. Conditioning on the
\emph{logger regime} it survives where it matters most --- $\rho=-0.56$ under an
independent logger, the regime that actually produces failures ($11.7\%$ of cells unsafe)
--- and weakens where there is almost nothing left to rank ($-0.27$ self-aligned and
$-0.17$ misaligned, with $2.8\%$ and $4.4\%$ unsafe cells). But conditioning on regime
\emph{and} $\tau$ jointly, the correlation collapses: median $-0.01$, negative in only
$6$ of $9$ strata. We do not read this as the diagnostic failing. Regime $\times$ $\tau$
\emph{is} the mechanism that generates overlap variation in this design, so conditioning
on both removes precisely the variation ESS exists to track, leaving residual
within-stratum noise. It does mean the honest claim is narrower than a
temperature-only design would suggest: ESS fraction ranks importance-weighting risk
\emph{across} the logging situations a practitioner might face, not \emph{within} a
single fixed one. That is still the practitioner-relevant statement --- nobody knows in
advance which regime their historical log came from --- but it is a claim about
between-situation ranking, and we make no stronger one.

\begin{figure}[t]
  \centering
  \includegraphics[width=\linewidth]{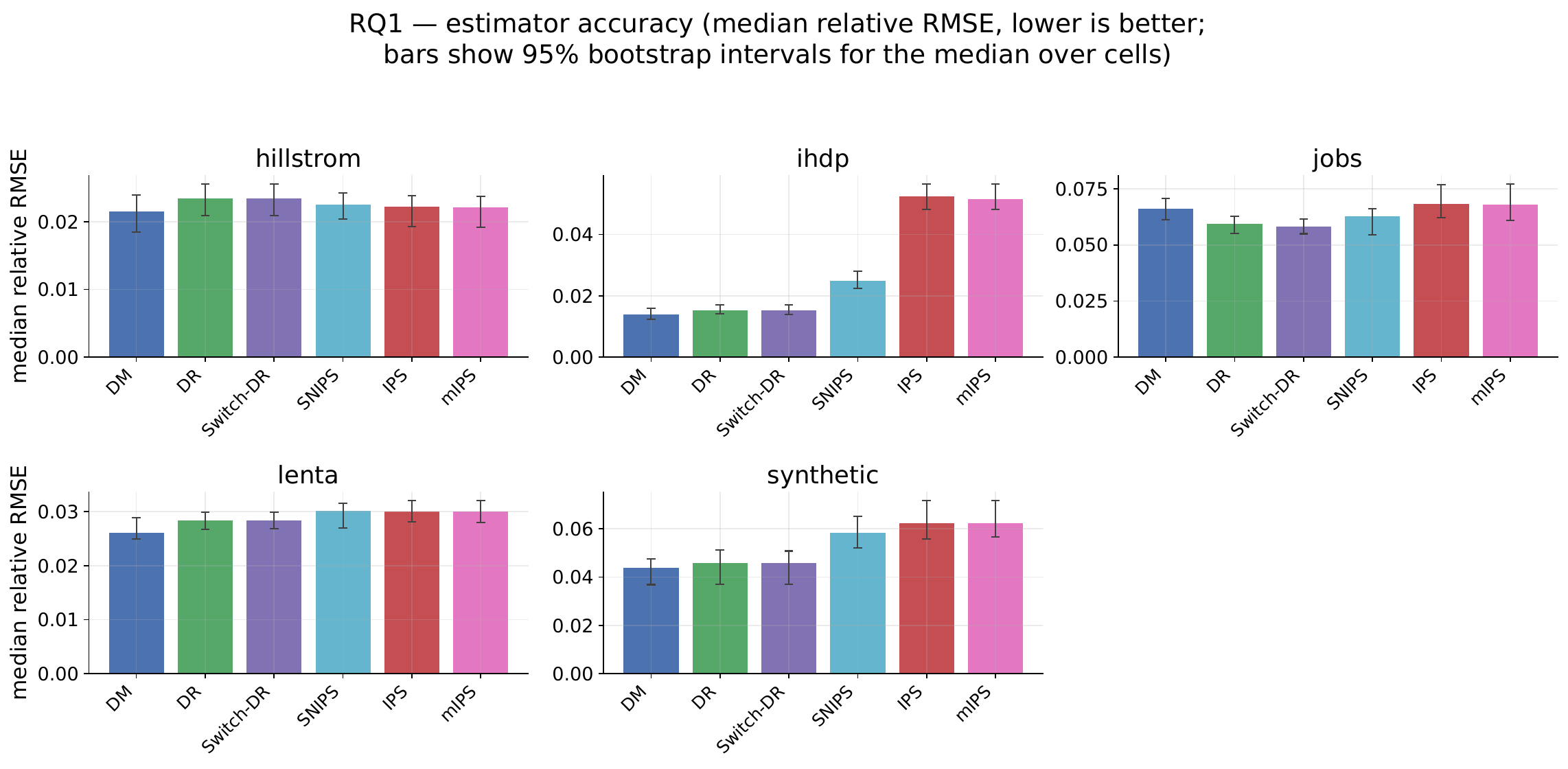}
  \caption{RQ1 --- median relative RMSE by estimator, per dataset (lower is better).
  \est{DM} and the DR family lead on four of five datasets; \est{IPS}/\est{mIPS} are
  worst, most dramatically on the continuous-outcome \dataset{IHDP}.}
  \label{fig:rq1}
\end{figure}

\begin{figure}[t]
  \centering
  \includegraphics[width=\linewidth]{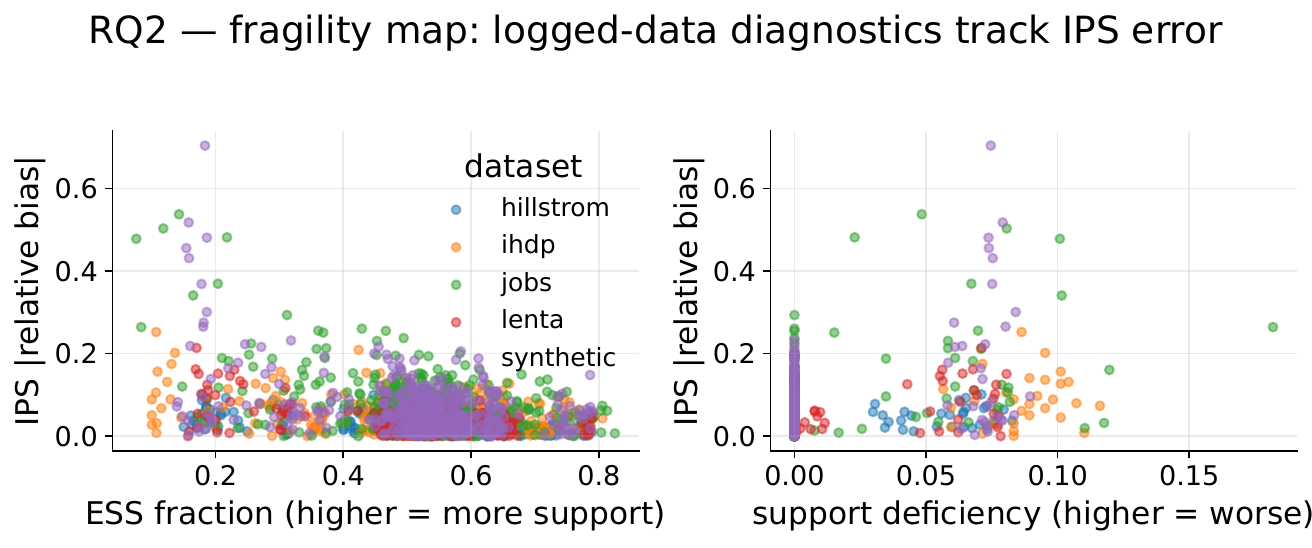}
  \caption{RQ2 fragility map --- IPS $|$relative bias$|$ against two logged-data
  diagnostics. Lower effective sample size (left) and higher support deficiency (right)
  both track larger error, quantified by the Spearman correlations in the text.}
  \label{fig:rq2-trust}
\end{figure}

\begin{table}[t]\centering
\caption{For \est{IPS}, propensity estimation is the dominant error source. Each row replaces the exact
$\pi_b$ with an out-of-fold estimate and re-runs the whole alignment sweep, with weights
\emph{and} diagnostics computed from the estimate. \textbf{Aggregation:} relative RMSE is
computed per cell (dataset $\times$ candidate $\times$ logger regime $\times$ overlap
$\times$ budget) as RMSE over the $10$ seeds divided by $|$mean true value$|$, then
\emph{averaged} over the $216$ exact-value cells (\dataset{IHDP}, \dataset{synthetic}) ---
a mean over cells, unlike Table~\ref{tab:rq1-accuracy}, which takes the median. Under the
median convention the levels are lower (exact control: \est{DM} $0.026$, \est{DR} $0.027$,
\est{IPS} $0.057$, \est{SNIPS} $0.040$) but every comparison below is unchanged. The
failure rate and AUC are over all $540$ cells. \est{DM} is constant by
construction (it uses no propensity) and serves as the pipeline check. \est{DR} is
essentially immune while \est{IPS} degrades several-fold, and the fragility screen
degrades from useful to inverted as $\hat\pi_b$ worsens.}
\label{tab:propensity}
\small
\begin{tabular}{llrrrrrr}
\toprule
$\hat\pi_b$ & quality ($\rho$ w/ truth) & \est{DM} & \est{DR} & \est{IPS} & \est{SNIPS} & \est{IPS} fail & ESS AUC \\
\midrule
exact (control)      & ---              & 0.036 & 0.037 & 0.072 & 0.052 & $6.3\%$  & $0.85$ \\
\est{LightGBM}($x$)  & $+0.30$          & 0.036 & 0.042 & 0.349 & 0.069 & $62.8\%$ & $0.84$ \\
logistic($x$)        & $+0.27$          & 0.036 & 0.036 & 0.137 & 0.054 & $46.3\%$ & $0.54$ \\
marginal (no $x$)    & $-0.02$          & 0.036 & 0.036 & 0.219 & 0.103 & $37.2\%$ & $\mathbf{0.05}$ \\
\bottomrule
\end{tabular}
\end{table}

\begin{figure}[t]
  \centering
  \includegraphics[width=\linewidth]{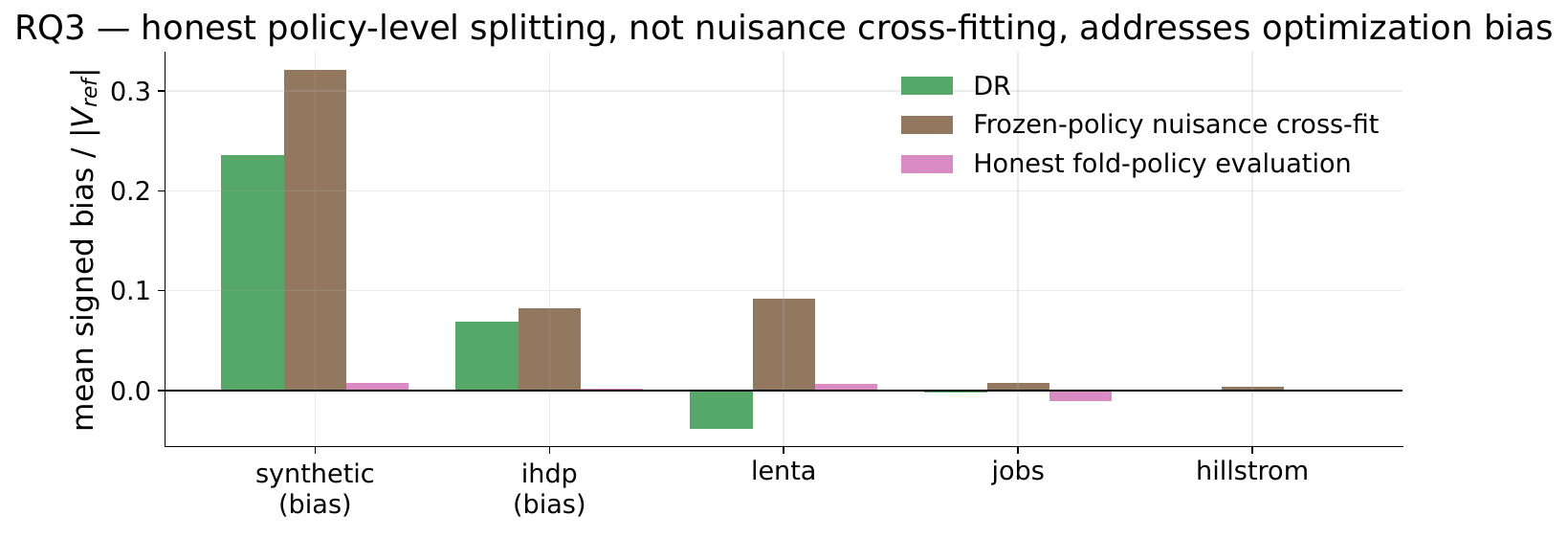}
  \caption{RQ3 --- honest policy-level splitting, not nuisance cross-fitting,
  addresses optimization bias. Mean signed bias under in-sample policy fitting, each
  estimator against its own estimand's reference. On the datasets with material
  detected bias (\dataset{IHDP}, \dataset{synthetic}) plain \est{DR} is strongly
  optimistic; frozen-policy nuisance cross-fit (same
  fixed-policy estimand) is \emph{more} optimistic, not less; honest fold-policy
  evaluation (learning-procedure estimand) is near zero.
  On the datasets with no material detected bias all three are near zero.}
  \label{fig:rq3}
\end{figure}

\FloatBarrier
\section{Formal estimator definitions}
\label{app:estimators}

\begin{table}[t]
\centering
\caption{Benchmark datasets: \emph{five primary} (\dataset{Synthetic}, \dataset{IHDP},
\dataset{Jobs}, \dataset{Hillstrom}, \dataset{Lenta}) plus \emph{two known-effect
hardening suites} (six ACIC-style DGPs each, over the \dataset{IHDP} and
\dataset{Hillstrom} covariates; Appendix~\ref{app:hardening}). $n$: sample size used
(large marketing RCTs capped at $50{,}000$ by uniform subsampling, which preserves the
constant propensity; raw size in parentheses). The \dataset{Twins} row is an external validation (Appendix~\ref{app:twins}), not a sixth primary dataset: no headline average includes it. $\pi_1$: treatment fraction. $\bar y$:
outcome mean (a base rate for binary outcomes; on the outcome scale for the continuous
\dataset{IHDP}). Reference: exact potential-outcome means where known (ground truth in
the strict sense), else an unbiased-but-noisy Horvitz--Thompson reference estimate on
the randomized split. For the hardening suites, $\pi_1$ describes the factual
confounded-logistic assignment (clipped to $[0.15,0.85]$); their logged feedback is
drawn from the known surfaces, so factual overlap is incidental there.}
\label{tab:datasets}
\small
\begin{tabular}{llrrrl}
\toprule
Dataset & Regime & $n$ & $\pi_1$ & $\bar y$ & Reference value \\
\midrule
\dataset{Synthetic} & Synthetic       & 7{,}000          & 0.51 & 0.51  & exact $\mu_0,\mu_1$ \\
\dataset{IHDP}      & Semi-synth.\ (cont.) & 672         & 0.18 & 3.17  & exact $\mu_0,\mu_1$ \\
\dataset{Jobs}      & Semi-synth.\ (RCT)   & 578         & 0.41 & 0.73  & HT reference estimate \\
\dataset{Hillstrom} & Marketing RCT   & 50{,}000 (64K)   & 0.67 & 0.15  & HT reference estimate \\
\dataset{Lenta}     & Marketing RCT   & 50{,}000 (687K)  & 0.75 & 0.11  & HT reference estimate \\
\midrule
\multicolumn{6}{l}{\emph{Hardening suites (Appendix~\ref{app:hardening}):}} \\
\dataset{ACIC-style} & Known-effect ($\times 6$) & 672 & 0.52 & 2.0--2.6 & exact $\mu_0,\mu_1$ \\
\dataset{ACIC-hs} & Known-effect ($\times 6$) & 10{,}000 & 0.50 & 2.1 & exact $\mu_0,\mu_1$ \\
\midrule
\multicolumn{6}{l}{\emph{External validation, not a primary dataset (Appendix~\ref{app:twins}):}} \\
\dataset{Twins} & Paired real outcomes & 11{,}400 & 0.50 & 0.83 & recorded $\mu_0,\mu_1$ (paired) \\
\bottomrule
\end{tabular}
\end{table}

\paragraph{Symbols.} Four symbol groups would otherwise be overloaded, so we fix them here.
$\tau$ is always the \emph{logging temperature}; the Switch-DR threshold is
$\lambda_{\mathrm{sw}}$ and the shrinkage-DR parameter $\lambda$. $\varepsilon=0.02$ is
always the \emph{propensity floor}; the support-deficiency threshold is written $\kappa$
(default $0.05$; $\delta$ is reserved for the \emph{error} target of Section~\ref{sec:rq2}) and the perturbation-DR noise scale $\sigma_p$. $M$ is always a
\emph{clipping cap}; the number of perturbation draws is $B_p=25$. The cohort-conditional
estimand of Section~\ref{sec:setup} is written $\V_n$, distinct from the population
$\V$.

\paragraph{Compute and software.} All experiments run on a single 10-core Apple M-series
CPU; the full reproduction chain (every \texttt{make} target in
Appendix~\ref{app:repro-targets}) takes roughly $17$ hours end to end, the longest single
target being the propensity sweep at about $3.3$ hours. Python $3.12$, \est{LightGBM},
scikit-learn and Hydra at the versions pinned in the released \texttt{pyproject.toml}.
Seeds run $42,\dots,51$ throughout, except the superseded
misaligned-logging check of Appendix~\ref{app:align}, which predates the ten-seed rerun; every result row carries its git commit hash. All six
estimators are our own implementations against a common interface, so that they share one
outcome model and one propensity input and differ only in the estimating equation; we do
not wrap \est{OBP} or \est{SCOPE-RL}, though we verify numerical agreement against the
former on identical inputs (\texttt{make reference-check}; Section~\ref{sec:design}).

\paragraph{Two terms used throughout.} A \emph{cell} is a seed-aggregated
dataset\,$\times$\,policy\,$\times$\,budget\,$\times$\,temperature tuple; a
\emph{configuration} is one seed-specific execution of a cell. \emph{Normalized regret}
of an estimator on a slate is
$\big(V(\pi^\star)-V(\hat\pi)\big)/\big(V(\pi^\star)-\min_j V(\pi_j)\big)$, where
$\hat\pi$ is the candidate the estimator ranks first, $\pi^\star$ the reference-best
candidate and the denominator the spread of the slate; it is $0$ for a correct pick and
$1$ for the worst, and is comparable across datasets whose value scales differ. In the
released parquets the three logger regimes carry the artifact identifiers
\texttt{self\_aligned}, \texttt{misaligned} and \texttt{independent}.

Logged data are tuples $(x_i, a_i, y_i)$ with logging propensity
$e_i = \pi_b(a_i \mid x_i)$; in this benchmark $e_i$ is \emph{known by
construction} (the logging policies are synthesized; see
Section~\ref{sec:design}), which favours importance-weighted estimators relative
to production settings where the propensity must be estimated. The target policy
contributes $\pi_i^e = \pi_e(a_i \mid x_i) \in \{0, 1\}$ for the deterministic
top-$k$ allocation, giving the importance weight $w_i = \pi_i^e / e_i$. The
outcome model $\hat\mu_a(x)$ is one LightGBM regressor per action, fit on the
logged sample and shared by all model-based estimators in a cell.

\paragraph{DM.} $\Vhat_{\mathrm{DM}} = \frac1n \sum_i
\hat\mu_{a_{\pi_e}(x_i)}(x_i)$.

\paragraph{IPS.} $\Vhat_{\mathrm{IPS}} = \frac1n \sum_i w_i\, y_i$
\citep{dudik2011doubly}. Unbiased given correct propensities and support; for a
deterministic target most $w_i$ are $0$ and the rest $1/e_i$.

\paragraph{SNIPS.} $\Vhat_{\mathrm{SNIPS}} = \sum_i w_i y_i \big/ \sum_i w_i$
\citep{swaminathan2015self}.

\paragraph{DR.} $\Vhat_{\mathrm{DR}} = \frac1n \sum_i \big[
\hat\mu_{a_{\pi_e}(x_i)}(x_i) + w_i\,(y_i - \hat\mu_{a_i}(x_i)) \big]$
\citep{dudik2011doubly}.

\paragraph{Switch-DR.} DR with the correction term dropped where $w_i > \lambda_{\mathrm{sw}}$ (we write the Switch-DR threshold as $\lambda_{\mathrm{sw}}$ to keep it distinct from the logging temperature $\tau$)
(falling back to DM there) \citep{wang2017optimal}:
$\Vhat_{\mathrm{sw}} = \frac1n \sum_i \big[ \hat\mu_{a_{\pi_e}(x_i)}(x_i) +
\mathbb{1}\{w_i \le \lambda_{\mathrm{sw}}\}\, w_i\,(y_i - \hat\mu_{a_i}(x_i)) \big]$.
$\lambda_{\mathrm{sw}}$ is selected \emph{on the logged sample} from the grid
$\{5, 10, 20, 50, 100, \infty\}$ by minimizing the standard estimated-MSE bound:
(dropped-correction magnitude)$^2$ $+$ sample variance of the retained
contributions.

\paragraph{Mixture-propensity IPS (\est{mIPS}).} IPS with the propensity replaced
by a mixture with uniform exploration,
$\Vhat_{\alpha} = \frac1n \sum_i \frac{\pi_i^e}{(1-\alpha)\,e_i +
\alpha/2}\, y_i$, keeping weights bounded (the mixed-logging device of
\citealp{nishimura2024coupon}). $\alpha$ is selected on the logged sample from
$\{0, 0.05, 0.1, 0.2, 0.5\}$ by the same estimated-MSE proxy (squared deviation
from the IPS point value as the bias proxy, plus sample variance). Replacing the true
propensity with the mixture denominator deliberately introduces bias in exchange for
bounded weights, so \est{mIPS} is a variance-control heuristic, not an unbiased
estimator of the behaviour policy's IPS; and because the bias proxy is anchored to the
raw-IPS point value, the tuning target is itself noisy under poor overlap (where IPS is
unstable) --- we therefore treat $\alpha$-selection as a heuristic, not a guarantee.

\paragraph{Perturbation-DR.} A smoothing heuristic inspired by
\citet{guo2022offpolicy}: with candidate scores $s$ and $B_p{=}25$ draws
$\xi_m \sim \mathcal{N}(0, \sigma_p^2 I)$, $\sigma_p = 0.5\,\mathrm{sd}(s)$, re-solve the
budgeted allocation $z_m = \mathrm{alloc}(s + \xi_m, k)$ and average the
DR values: $\Vhat_{\mathrm{pert}} = \frac1{B_p} \sum_m \Vhat_{\mathrm{DR}}(z_m)$.
Its estimand is therefore the expected value of a \emph{perturbation-smoothed
policy distribution} around the deterministic allocation --- not the
deterministic policy itself --- which is why it is excluded from the fixed-policy
RQ1, RQ3, and RQ4 comparisons and discussed separately in
Appendix~\ref{sec:smoothing}.

\paragraph{Cross-fitted DR (RQ3 only).} Two modes with distinct estimands
(Section~\ref{sec:rq3}): \emph{frozen-policy} (the policy is fixed; only
$\hat\mu$ is fit out-of-fold, $5$ folds --- same estimand as DR) and
\emph{fold-policy} (score and allocation re-derived per fold from the
out-of-fold model --- estimand is the value of the learning algorithm, scored
against its own fold-matched exact reference).

\paragraph{Aggregation.} Let an experimental \emph{configuration} $c$ be a
(dataset, candidate policy, budget, overlap temperature) tuple, evaluated over seeds
$s = 1,\dots,S$; for estimator $e$ write its estimate as $\Vhat_{e,c,s}$ and the
reference value as $V_{c,s}$ (the reference does not depend on $e$). Define
\[
  \mathrm{relRMSE}_{e,c} \;=\;
  \frac{\sqrt{\tfrac1S \sum_{s} (\Vhat_{e,c,s} - V_{c,s})^2}}
       {\big|\tfrac1S \sum_{s} V_{c,s}\big|},
\]
i.e.\ RMSE over seeds normalized by the configuration's absolute mean reference value
(configurations with near-zero mean reference would be dropped rather than inflated;
none occur in the released runs). The table entry for a (dataset, estimator) cell is
then $\mathrm{median}_c\, \mathrm{relRMSE}_{e,c}$ over the configurations in that cell ---
the median, because IPS-family errors are heavy-tailed under poor overlap. We spell
this out because the order (RMSE over seeds, then normalize, then median over
configurations) is not the only reading of ``median relative RMSE.'' The reported
bootstrap CIs are \emph{conditional} contribution-level intervals: they resample the
completed per-unit DR/IPS contributions on the evaluation split and do not refit the
nuisance models, so they reflect sampling variation given the fitted $\hat\mu$, not the
uncertainty of fitting it.

% ============================================================
\section{Policy-smoothing sensitivity (perturbation-DR)}
\label{sec:smoothing}

Perturbation-DR changes the target from the fixed deterministic allocation $V(\pi_e)$
to the expected value of a distribution of perturbed allocations,
$\mathbb{E}_{\varepsilon}[V(\pi_{s+\varepsilon})]$ (Appendix~\ref{app:estimators}). It
therefore cannot be ranked directly against the fixed-policy estimators in RQ1, the
fixed-policy optimization-bias estimators in RQ3, or across an RQ4 candidate slate
(it perturbs each candidate's own scores, so no two candidates share an estimand). The estimand-coherent evaluation
scores it against the averaged exact values of its own perturbed policies,
$V_{\mathrm{pert,ref}} = \frac1M\sum_m V_{\mathrm{exact}}(z_m)$, where the $z_m$
replicate the estimator's own perturbation draws (same seed and noise stream, so both
sides average the same $M{=}25$ allocations). We ran this on the exact-value datasets
over the main grid (\dataset{Synthetic} and \dataset{IHDP}, both candidate learners,
$\tau\in\{0.5,2,5\}$, the five non-trivial budgets --- at $k{=}1$ perturbation cannot
change the allocation --- ten seeds; $60$ cells): median $|$relative error$|$
is $1.6\%$ ($1.9\%/1.4\%/1.4\%$ at $\tau=0.5/2/5$), and no cell exceeds $10\%$.
Perturbation-DR is therefore accurate \emph{on its own estimand}; its exclusion from
the fixed-policy tables reflects an estimand gap, not an estimation failure. Because it
evaluates a different deployment policy (a smoothed allocation, which practitioners may
or may not wish to deploy), it is still reported separately
(\texttt{make repro-perturbation-matched}).

% ============================================================
\section{Retained sample size by overlap regime}
\label{app:retained-n}

Rejection sampling retains fewer units under sharper logging, so the overlap knob
moves both propensity concentration and logged sample size. Median retained $n$ per
temperature:

\begin{center}
\small
\begin{tabular}{lrrr}
\toprule
dataset & $\tau{=}0.5$ (poor) & $\tau{=}2.0$ & $\tau{=}5.0$ (good) \\
\midrule
\dataset{Hillstrom} & $8{,}470$ & $9{,}225$ & $11{,}733$ \\
\dataset{Lenta}     & $6{,}347$ & $6{,}752$ & $8{,}597$ \\
\dataset{Synthetic} & $1{,}807$ & $2{,}091$ & $2{,}631$ \\
\dataset{Jobs}      & $132$     & $168$     & $203$ \\
\dataset{IHDP}      & $336$     & $336$     & $336$ \\
\bottomrule
\end{tabular}
\end{center}

Poor overlap therefore also costs $\sim\!25$--$35\%$ of the logged sample on the
rejection-sampled datasets. The two effects can be separated using \dataset{IHDP},
whose surface-sampling construction retains \emph{every} unit at every temperature
($n$ constant at $336$): there the temperature sweep changes only the propensity
concentration, and the IPS-family degradation under a misaligned logger persists
(Figure~\ref{fig:rq2-alignment}) while the DR family stays flat.
Propensity concentration, not the incidental sample-size reduction, is the dominant
mechanism; a fully matched-$n$ control on the rejection-sampled datasets is left to
future work.

% ============================================================
\section{Cluster-bootstrap detail for the RQ2 diagnostics}
\label{app:cluster}

Complementing the within-dataset and leave-one-dataset-out analysis in
Section~\ref{sec:rq2}, a percentile bootstrap of the Spearman $\rho$ between each
diagnostic and IPS $|$relative bias$|$, resampling whole clusters ($4{,}000$
replicates), gives the following 95\% intervals. The statistic is computed over
\emph{cells} (Section~\ref{sec:design}), which already aggregate the ten seeds, so seeds
are not an available resampling level; the two levels we can resample are the dataset and
the (dataset, candidate policy) block, of which there are $5$ and $10$ respectively on the
alignment sweep (two learned candidates per dataset):

\begin{center}
\small
\begin{tabular}{lcc}
\toprule
diagnostic & dataset clusters ($5$) & dataset$\times$policy clusters ($10$) \\
\midrule
ESS fraction        & $[-0.56,\,-0.33]$ & $[-0.53,\,-0.33]$ \\
support deficiency  & $[+0.39,\,+0.60]$ & $[+0.34,\,+0.54]$ \\
maximum weight      & $[+0.26,\,+0.59]$ & $[+0.25,\,+0.52]$ \\
\bottomrule
\end{tabular}
\end{center}

All intervals exclude zero, and the finer (dataset\,$\times$\,policy) clustering gives the
narrower intervals, as one would expect from more resampling units. We stress that the
narrowness of the five-cluster intervals is
not evidence of a large effective sample: with only five top-level clusters, it reflects how similar the
within-dataset correlations are (Section~\ref{sec:rq2}). We therefore treat the
within-dataset consistency and leave-one-dataset-out stability, not these intervals, as
the primary evidence.

% ============================================================
\section{Paired comparisons and interval coverage}
\label{app:paired}

\subsection{Interval coverage against exact references}
\label{app:coverage}

The title's question is about trust, and trust in an interval is coverage. Every
estimator reports a $95\%$ percentile bootstrap CI over per-unit contributions
(conditional on fitted nuisances); on the exact-value datasets of the alignment sweep we
can score those intervals against the truth. Coverage confirms the two warnings the paper
states about its own intervals, and localizes them:

\begin{center}\small
\begin{tabular}{lrrrr}
\toprule
 & pooled & self-aligned & misaligned & independent \\
\midrule
\est{DM}        & $0.89$ & $0.91$ & $0.90$ & $0.87$ \\
\est{DR}        & $0.95$ & $0.96$ & $0.96$ & $0.92$ \\
\est{IPS}       & $0.96$ & $0.97$ & $0.96$ & $0.93$ \\
\est{SNIPS}     & $0.98$ & $0.99$ & $0.99$ & $0.96$ \\
\bottomrule
\end{tabular}
\end{center}

\est{DM} undercovers everywhere ($0.89$ pooled, $0.87$ on \dataset{synthetic}): its
conditional intervals ignore outcome-model refit variance, exactly the deficiency
Section~\ref{sec:cannot} flags, and the reason we call the model-based intervals
optimistic. \est{IPS} covers near-nominally where weighting is safe but degrades to
$0.89$ under the independent logger at $\tau{=}0.5$ --- coverage collapses in the same
cells the fragility screen flags, so the screen also marks where the \emph{intervals}
stop being trustworthy. \est{SNIPS} overcovers because its intervals are wide (median
width $0.20|V|$ against $0.12|V|$ for \est{DM}). None of these are population-coverage
statements; they are conditional-bootstrap coverage on this benchmark's exact-value
cells ($12{,}960$ estimator-configuration intervals; $2{,}160$ per estimator).

\textbf{Mean versus median paired differences (RQ1).} The
$\est{IPS}-\est{DR}$ gap of $+0.015$ ($+0.013$ on the $k<1$ grid) in
Section~\ref{sec:rq1real} is a \emph{mean} paired difference over cells; the
corresponding \emph{median} is $+0.007$ on both grids. The mean is
larger precisely \emph{because} it is a mean: the gap between the two summaries is the
right tail of \est{IPS} relative RMSE under a misaligned logger --- a minority of cells in
which weighting fails badly, which a mean registers and a median does not. Both agree in
sign and ordering; we report the mean because that tail is the practically important part
of the distribution, and flag that it is tail-driven rather than typical.

\textbf{Within-family comparisons (RQ1).} \est{Switch-DR} is statistically
indistinguishable from \est{DR} ($-0.0002$, CI $[-0.0007,+0.0001]$). The
\est{DM}-over-\est{DR} margin is at most \emph{tiny}: $\est{DR}-\est{DM}=+0.0007$
(CI $[-0.0012,+0.0021]$, which includes zero; sign convention throughout is first-minus-second, so positive here
means \est{DM} is better; \est{DM} better in only $61\%$ of cells) --- and this interval is
\emph{conditional} (it resamples completed contributions without refitting the outcome
model; Appendix~\ref{app:estimators}), so it understates uncertainty for the model-based
estimators. Under a full nuisance-refit resampling (\DMDRREFIT), the
\est{DM}--\est{DR} difference is not distinguishable from zero. We therefore do
\emph{not} claim a separation of \est{DM} from \est{DR}: the robust, practically
meaningful claim is the model-based \emph{family} advantage over importance weighting.

\section{Logging construction: properties and retained sample size}
\label{app:logging}

The rejection sampler of Section~\ref{sec:design} has two properties that bear on how the
overlap sweep should be read.

\textbf{Centring.} Standardizing $s$ before the logistic matters for comparability: raw
score scales differ across learners, so without it a common $\tau$ would mean different
degrees of logging sharpness for different candidates. But the logistic is centred at
the score \emph{mean}, not at the budget-dependent top-$k$ cutoff $q_{1-k}$, so the
logger's marginal treatment rate is $\approx\!0.5$ regardless of $k$. A given $\tau$
therefore does not correspond to the same overlap for a top-$10\%$ and a top-$70\%$
target: $\tau$ is comparable across learners but not across budgets. A budget-centred
logger $\sigma((s-q_{1-k})/\tau)$ removes this, and we ran it (Appendix~\ref{app:cutoff}).

\textbf{Retained sample size co-varies with overlap.} Rejection sampling retains fewer
units at low $\tau$, so the overlap sweep moves retained $n$ as well as propensity
concentration (Appendix~\ref{app:retained-n} tabulates both). \dataset{IHDP}, whose
logged sample is drawn from known surfaces at constant $n$, is the control: the
IPS-error inflation appears there too, so it is not purely a sample-size effect. We note that the differential pattern --- \est{IPS} inflating while the DR family stays
flat --- is \emph{not} itself evidence against the confound, since \est{IPS} error is
variance-dominated ($\propto\mathbb{E}[w^2]/n$) and would inflate under a shared
$n$-loss while a bias-dominated \est{DM} would not. The constant-$n$ \dataset{IHDP}
control is what carries the argument.
A matched-$n$ control on the rejection-sampled datasets is left to future work.

\section{Robustness checks: full results}
\label{app:robustness}

Full tables and mechanism discussion for the outcome-model-quality ladder of
Section~\ref{sec:omquality} and the two robustness checks of
Section~\ref{sec:robustness}.

\subsection{Per-dataset accuracy under out-of-fold nuisances}
\label{app:oof-perdataset}
Table~\ref{tab:rq1-accuracy} reports the out-of-fold-nuisance rerun only as two
cross-dataset means. Table~\ref{tab:rq1-oof} gives the per-dataset detail behind them, so
that every per-dataset claim about out-of-fold nuisances --- including item~2 of
Section~\ref{sec:guide} --- can be checked against the paper itself.
\begin{table}[t]\centering
\caption{Per-dataset accuracy under \emph{out-of-fold} nuisances (\texttt{make repro-full-oof}), the per-dataset detail behind the two out-of-fold mean columns of Table~\ref{tab:rq1-accuracy}. Median relative RMSE, same convention as Table~\ref{tab:rq1-accuracy}; best per column in bold. Four decimals because several columns separate only in the fourth. \est{IPS}, \est{SNIPS} and \est{mIPS} use no outcome model and are numerically identical to their in-sample counterparts. \est{DM} is best or tied-best on three of the five datasets here (\dataset{Hillstrom} within $0.0001$ of \est{mIPS}, \dataset{IHDP}, \dataset{synthetic}) against four of five in-sample, and is the \emph{worst} of the six on \dataset{Jobs} and \dataset{Lenta} --- both HT-reference datasets, where Section~\ref{sec:rq1} argues the ordering is not resolvable.}
\label{tab:rq1-oof}
\small
\begin{tabular}{lrrrrrrr}
\toprule
estimator & hillstrom & ihdp & jobs & lenta & synthetic & mean (exact) & mean (RCT-ref) \\
\midrule
dm & 0.0222 & \textbf{0.0153} & 0.0704 & 0.0318 & \textbf{0.0456} & \textbf{0.0304} & 0.0414 \\
dr & 0.0244 & 0.0166 & 0.0600 & 0.0310 & 0.0467 & 0.0317 & 0.0384 \\
switch\_dr & 0.0244 & 0.0166 & \textbf{0.0588} & 0.0310 & 0.0467 & 0.0317 & \textbf{0.0381} \\
snips & 0.0225 & 0.0248 & 0.0626 & 0.0302 & 0.0582 & 0.0415 & 0.0384 \\
ips & 0.0222 & 0.0523 & 0.0683 & \textbf{0.0300} & 0.0623 & 0.0573 & 0.0402 \\
mIPS & \textbf{0.0221} & 0.0515 & 0.0681 & \textbf{0.0300} & 0.0623 & 0.0569 & 0.0401 \\
\bottomrule
\end{tabular}
\end{table}

\subsection{Known-effect hardening on real covariates}
\label{app:hardening}

A skeptical reading of the evidence so far is that the exact-value evidence rests on
one synthetic generator and one semi-synthetic dataset. The hardening sweep addresses
this with six additional known-effect DGPs on the \emph{real} \dataset{IHDP}
covariates (Section~\ref{sec:design}), where the covariate joint distribution is real
and the truth is exact by construction.
\begin{table}[t]\centering
\caption{Known-effect hardening: median relative RMSE on six ACIC-2017-style DGP settings over the real \dataset{IHDP} covariates (S1/S2 linear, S3/S4 nonlinear, S5/S6 step-subgroup surfaces; odd $=$ low noise, even $=$ high). Best per column in bold.}
\label{tab:acic-hardening}
\begin{tabular}{lrrrrrrr}
\toprule
estimator & S1 & S2 & S3 & S4 & S5 & S6 & mean (exact) \\
\midrule
dm & \textbf{0.019} & \textbf{0.073} & \textbf{0.013} & \textbf{0.071} & \textbf{0.016} & \textbf{0.026} & \textbf{0.036} \\
dr & 0.023 & 0.090 & 0.015 & 0.081 & \textbf{0.016} & 0.029 & 0.042 \\
switch\_dr & 0.023 & 0.090 & 0.015 & 0.081 & \textbf{0.016} & 0.029 & 0.042 \\
snips & 0.036 & 0.099 & 0.019 & 0.087 & 0.025 & 0.035 & 0.050 \\
ips & 0.061 & 0.110 & 0.057 & 0.102 & 0.057 & 0.062 & 0.075 \\
mIPS & 0.061 & 0.108 & 0.057 & 0.099 & 0.057 & 0.061 & 0.074 \\
\bottomrule
\end{tabular}
\end{table}
Table~\ref{tab:acic-hardening} shows the RQ1 ordering replicates in full: \est{DM} is
the most accurate (mean relative RMSE $0.036$, and lowest in \emph{every one} of the six
settings --- its closest call is S1, $0.019$ against \est{Switch-DR}'s $0.023$), the DR
family follows ($0.042$), and
\est{IPS}/\est{mIPS} are worst \emph{in every setting} ($0.075$ and $0.074$ mean,
$2.1\times$ \est{DM}). The margin is largest on the
low-noise settings, where the outcome model is easy to fit; high noise (S2, S4)
compresses all estimators toward the same error, exactly as the outcome-model-quality
caveat of Section~\ref{sec:omquality} predicts. The fragility map of Section~\ref{sec:rq2} also
replicates independently on this sweep: Spearman $\rho$ between IPS $|$relative
bias$|$ and ESS fraction / support deficiency / max weight is
$-0.37$ / $+0.29$ / $+0.23$ on this temperature-design sweep --- distinct from the
alignment-axis rerun of the same DGPs quoted in Section~\ref{sec:robustness}, which gives
$-0.45$ --- against $-0.41$ / $+0.42$ / $+0.35$ on the primary alignment sweep; same
signs, comparable magnitudes (the fragility-screen thresholds are validated on the alignment-axis rerun of these
DGPs, not on this sweep; cf.
Section~\ref{sec:rq2}).

Because the IHDP covariate matrix is small ($n{=}672$), we additionally repeat the
six-DGP sweep on a \emph{second, independent} real covariate set: a fixed
$10{,}000$-row subsample of the \dataset{Hillstrom} marketing covariates (another
$2{,}160$ configurations, zero anomalies). All conclusions replicate: \est{DM} is most
accurate (mean relative RMSE $0.0055$, and lowest in every one of the six settings), the
DR family follows ($0.0062$), \est{IPS}/\est{mIPS} are worst in every setting ($0.0176$
and $0.0177$, $3.2\times$ \est{DM}), and the diagnostics correlate with error comparably
($\rho = -0.52$ / $+0.34$ / $+0.37$ on this sweep). Absolute errors are roughly an order of
magnitude smaller than on the $n{=}672$ IHDP covariates at comparable overlap ---
a reminder that logged sample size and overlap both drive error, and the benchmark's
relative comparisons are the stable object, not the absolute error levels. Exact ground truth built on real
covariates therefore tells the same story as both the fully synthetic and the
RCT-reference evidence.

\subsection{External validation on a non-simulated paired reference}
\label{app:twins}

The hardening sweeps answer one version of the skeptical reading. A sharper version
survives them: every exact-value surface in this paper is a \emph{simulated} response
surface --- our own synthetic generator and ACIC-style constructions, and
\dataset{IHDP}'s canonical simulated surfaces, which we adopt rather than build --- so
an estimator is ultimately scored against outcomes somebody modelled. This appendix
removes that reliance using a dataset whose evaluation reference is built directly from
recorded paired outcomes rather than from any modelled surface.

\paragraph{Design.} The \dataset{Twins} cohort \citep{almond2005costs}, in the form
distributed with \citet{yoon2018ganite}, contains $11{,}400$ same-sex twin pairs born
under $2$\,kg with $30$ covariates. Taking the \emph{pair} as the unit, the treatment is
being the heavier twin and the outcome is survival through the first year. Both
siblings' outcomes are recorded, so the pair supplies two \emph{recorded} outcome
coordinates and $\mu_0,\mu_1$ are read off the data. Be precise about what this buys.
These are not two potential outcomes observed for one infant: the twins are different
individuals, so this is a non-simulated \emph{paired} reference, and it proxies a causal effect of
birth weight only under co-twin exchangeability given the shared covariates --- an
assumption the design makes plausible but does not prove. The narrower property is the
one we need and it does hold: the reference value is fixed by the data and not by any
surface we chose, so scoring against it is not circular. Marginal survival is $0.823$
for the lighter twin against $0.839$ for the heavier --- a paired heavier-minus-lighter
survival difference of $+1.61$pp, reproducing the published figures for this cohort. Because the two paired reference outcomes are recorded binary values and the factual
outcome is set to the corresponding recorded one, there is no response-surface sampling
noise here (the calibrated noise of Appendix~\ref{app:estimators} is exactly zero):
logged rewards are actual recorded outcomes and the retained comparison outcome is the
actual co-twin outcome. The
protocol is otherwise unchanged (\texttt{make repro-twins}).

\paragraph{The mechanisms replicate; two of them more strongly.} The RQ1 ordering holds
exactly --- \est{DM} $0.0031$, the DR family $0.0035$, \est{SNIPS} $0.0050$,
\est{IPS}/\est{mIPS} $0.0157$ --- and the model-based margin is \emph{larger} than on
simulated exact-value data: $5.0\times$ against the $2.0\times$ of
Table~\ref{tab:rq1-accuracy}. The RQ2 mechanism replicates in the form
Section~\ref{sec:rq2} states it: sharpening a self-aligned logger leaves overlap
flat (median ESS fraction $0.604$, $0.556$, $0.527$ at $\tau=0.5,2,5$), while the
misaligned and candidate-independent loggers collapse under the same sharpening
($0.388$ and $0.173$ at $\tau{=}0.5$). The screen's association is comparable to the
primary sweep: Spearman $\rho$ between IPS $|$relative bias$|$ and ESS fraction /
support deficiency / max weight is $-0.51$ / $+0.51$ / $+0.36$, against
$-0.41$ / $+0.42$ / $+0.35$, and its per-regime ordering is
preserved, weakest under the misaligned logger ($-0.16$) and strongest under the
independent one ($-0.76$). RQ3 replicates as well: plain DR is optimistic by $6.0\%$ of
$|V_{\mathrm{ref}}|$ --- materially so by the criterion of Table~\ref{tab:rq3-debias},
and close to \dataset{IHDP}'s $+6.9\%$ --- nuisance-only cross-fitting makes it
\emph{worse} ($-25.4\%$), inside the $-16\%$ to $-36\%$ band of
Section~\ref{sec:rq3}, and honest policy-level splitting reduces fold-matched bias by
$94.7\%$ on the learning-procedure estimand. We
report that last figure alongside the $58$--$92\%$ range rather than inside it: those
eight regimes come from two simulated generator families, and \dataset{Twins} is
evidence of a different kind.

\paragraph{What does not transfer is the calibration.} Two of the paper's numbers are
simply inoperative here, and both are ones Section~\ref{sec:cannot} already flags as
benchmark-specific. The $10\%$ relative-bias failure criterion is never reached --- the
largest IPS $|$relative bias$|$ anywhere in the run is $0.095$ --- so all three logger
regimes record a $0.0\%$ failure rate, and the headline triple of
Section~\ref{sec:rq2} has no analogue. Relative RMSE compresses by roughly an order of
magnitude, because it normalizes by $|V| \approx 0.83$ while the spread between the best
and worst candidate policy is about $2$pp; this is the gross-value metric's documented
tendency to reward baseline predictability, in its most extreme form. Read on the
policy-relevant scale the compression reverses the impression: \est{IPS}'s error is
$62\%$ of the entire achievable spread it would need to resolve, against $13\%$ for
\est{DM} and the DR family. (We did not re-run the RQ4 selection protocol here, so this
is a statement about error scale, not a measured selection result.) The lesson we draw is the one
Section~\ref{sec:cannot} states in advance: the \emph{mechanisms} transfer to a
non-simulated reference we did not construct, and the \emph{cut points} do not.

\subsection{Outcome-model quality: the other axis of the decision map}
\label{app:misspec}

Section~\ref{sec:omquality} states the finding; this appendix gives the design and the
per-DGP detail. We re-run the protocol on the exact-truth datasets with structured surfaces (the nonlinear synthetic generator, ACIC
S3 nonlinear, and ACIC S5 step) while degrading
the shared $\hat\mu$ along a quality ladder: LightGBM (well-specified), depth-2
stumps (weak), and ridge regression (misspecified: linear on non-linear/step surfaces);
the candidate policies are unchanged, so only the \emph{estimators'} model quality
moves. Table~\ref{tab:misspec} reports median relative RMSE \emph{per DGP} --- pooling
across DGPs would average away the effect, since the ACIC surfaces are far more
learnable than the synthetic one.
\begin{table}[t]\centering
\caption{Outcome-model misspecification, \emph{per DGP}: median relative RMSE by DGP and outcome-model quality (pooled over overlap and budgets). ``best'' ranges over the four shown estimators. The inversion is DGP-specific: on the nonlinear \dataset{synthetic} surface a degraded $\hat\mu$ makes \est{DM} the worst of the four; on the smoother ACIC surfaces \est{DM} stays competitive. On the ACIC step surface the misspecified linear model slightly \emph{beats} \est{LightGBM}, because the arm means there are nearly linear, so the flexible learner buys variance without reducing bias.}
\label{tab:misspec}
\begin{tabular}{llrrrrl}
\toprule
DGP & $\hat\mu$ quality & DM & DR & SNIPS & IPS & best \\
\midrule
synthetic (nonlin.) & strong (LightGBM) & 0.036 & 0.037 & 0.049 & 0.058 & dm \\
 & weak (stumps) & 0.099 & 0.049 & 0.049 & 0.058 & snips \\
 & linear (missp.) & 0.061 & 0.043 & 0.049 & 0.058 & dr \\
\midrule
ACIC nonlin. & strong (LightGBM) & 0.011 & 0.014 & 0.017 & 0.056 & dm \\
 & weak (stumps) & 0.011 & 0.015 & 0.017 & 0.056 & dm \\
 & linear (missp.) & 0.012 & 0.015 & 0.017 & 0.056 & dm \\
\midrule
ACIC step & strong (LightGBM) & 0.014 & 0.015 & 0.022 & 0.054 & dm \\
 & weak (stumps) & 0.016 & 0.018 & 0.022 & 0.054 & dm \\
 & linear (missp.) & 0.006 & 0.006 & 0.022 & 0.054 & dr \\
\bottomrule
\end{tabular}
\end{table}
Per-DGP detail: on the nonlinear \dataset{synthetic} surface \est{DM}'s median relative
RMSE goes $0.036 \rightarrow 0.099$ under stumps, overtaking even \est{IPS} ($0.058$),
and sits at $0.061$ under ridge; \est{DR} degrades far more gracefully
($0.037 \rightarrow 0.049$ / $0.043$), its correction term absorbing much of the model
error --- though not quite best at every rung there: \est{DR} is the only model-based estimator that never becomes worst. The Table~\ref{tab:misspec}
entries are per-DGP medians over the sweep. The headline crossing was subsequently
seed-resampled (Section~\ref{sec:omquality}): at the stump rung on \dataset{synthetic}
the \est{DM}$-$\est{IPS} gap is $[+0.022,\,+0.063]$, \est{DM} worse in $100\%$ of
$2{,}000$ seed resamples, so the $0.099$-vs-$0.058$ ordering is a resampling-tested
separation there. The remaining rungs and the ACIC families carry point estimates only,
and for those we still read the qualitative crossing rather than the precise gaps. On the smoother ACIC
nonlinear and step surfaces, the degradation is mild --- \est{DM} remains best or
within a whisker of \est{DR} even under the misspecified linear model --- because a
linear/stump fit still approximates those surfaces adequately.

A natural question is whether model-based fragility admits a validated screen the way
importance-weighting fragility does (RQ2). We attempted the obvious candidate:
an out-of-fold held-out RMSE of $\hat\mu$ on the logged sample, per arm and pooled,
normalized by the logged-reward standard deviation --- computable without ground
truth, recorded for every cell of this sweep. The honest answer is that it does
\emph{not} transfer: pooled across the misspecification grid, its Spearman
correlation with realized \est{DM} error is $\approx\!-0.10$ (flagging AUC $0.48$
--- uninformative), even though it carries real signal \emph{within} a DGP
($\rho$ up to $+0.42$ on the nonlinear synthetic ladder). The mechanism is
instructive: held-out predictive error conflates irreducible outcome noise with
structural misfit, and a high-noise DGP scores ``poor model'' while \est{DM}'s
estimate of the \emph{mean} remains accurate. Weighting fragility admits a useful observable risk ranking because its failure
mechanism is encoded in the importance weights themselves; in our study, model-based
fragility does \emph{not} admit an analogous transferable screen from ordinary factual
prediction error. Constructing a validated, transferable model-adequacy screen for OPE
--- the \est{DM}/\est{DR} analogue of ESS --- is therefore an open problem, and the
qualitative checks below are prudent checks supported by our analysis, not a
substitute for a known quantitative rule. The lesson is not that
\est{DM} generically collapses, but that model-based estimators inherit whatever error
the outcome model makes, and that error is DGP-dependent. When overlap is good but the surface may defeat the model,
\est{DR}'s doubly-robust correction makes it the safer default than pure \est{DM};
when the model is trustworthy, \est{DM}/\est{DR} lead; when both overlap and model are
poor, no estimator here is dependable. Crucially, outcome-model quality for
\emph{policy evaluation} is not fully certified by a single global held-out residual
(a model can predict factual outcomes well yet rank treatment effects poorly, or lack
support in one arm over part of the covariate space); a practitioner should check
held-out error \emph{per treatment arm}, calibration across the score distribution,
and sensitivity across outcome-model classes.

\subsection{Misaligned logging}
\label{app:align}

The main sweep uses \emph{self-aligned} logging (Section~\ref{sec:design}): the logger
is constructed from the same candidate score used by the evaluated policy --- not from
its top-$k$ action boundary --- which cleanly isolates
the overlap axis but does not probe the case where logging and target disagree. Since
OPE is often requested precisely when a new candidate differs from the deployed logger,
we cross the two: for the \est{T-} and \est{S-learner} candidates we log under one and
evaluate the other (\emph{misaligned}), and compare to the aligned case (the logger built from the evaluated candidate's own
score), holding everything else fixed.
\ALIGNRESULT

\section{RQ4: reference-type split and the superseded per-candidate design}
\label{app:rq4-percand}

A composition note first, since slate composition is one of this section's own lessons.
The competitive slate holds seven candidates on the three RCT datasets but six on
\dataset{Synthetic} and \dataset{IHDP}: the class-transformation learner requires a
binary outcome and both exact-value datasets are continuous, so its failure is
deterministic, and within every cell all estimators rank the identical surviving slate.
Because the six-candidate cells are \emph{exactly} the exact-value datasets, slate size
is perfectly confounded with reference type here --- restricting to all-seven cells just
reproduces the HT-reference subset, whose reversal of the pooled gap is the
reference-type split reported in Appendix~\ref{app:refdep}, not a slate-size effect.

These are the full selection tables for the \emph{per-candidate} (self-aligned) logging
design, in which each candidate is scored on a log built from its own score. As
Section~\ref{sec:rq4} shows, that design compares policy--logger pairs rather than
policies; the
common-logger results of Table~\ref{tab:rq4-logger} are the primary evidence. We retain
these for completeness and because the contrast between the two designs is itself the
finding. Note in particular that the ``\est{IPS} selects at or below the uniform
baseline'' reading suggested by these numbers does \emph{not} survive the shared-log
design.

Correct-selection rate under the common \emph{independent} logger, split by reference
type (the numbers behind the reference-split discussion in Section~\ref{sec:rq4}):
\begin{center}\small
\begin{tabular}{llrr}
\toprule
slate & estimator & exact-value & HT-reference \\
\midrule
\multirow{2}{*}{3 cand.} & \est{DM}  & 0.883 & 0.513 \\
                         & \est{IPS} & 0.620 & 0.596 \\
\midrule
\multirow{2}{*}{7 cand.} & \est{DM}  & 0.550 & 0.196 \\
                         & \est{IPS} & 0.239 & 0.378 \\
\bottomrule
\end{tabular}
\end{center}

\begin{table}[t]\centering
\caption{RQ4 decoy over-selection: how often each estimator selects the random-score
candidate, against the rate at which that candidate is genuinely reference-best. A random
score induces smooth, well-overlapped logging, so an estimator evaluates it unusually
stably and may prefer it for reasons unrelated to its value --- most strongly for
weighting, but \est{DM} shows the same preference at roughly half the strength under the
two aligned designs, matching or exceeding \est{IPS} under the common-independent one
($16.3\%$ vs $16.8\%$; $8.7\%$ vs $8.2\%$). The effect therefore varies with the logging
design rather than being a fixed property of the estimators.}
\label{tab:rq4-decoy}
\small
\begin{tabular}{llrrr}
\toprule
slate & estimator & per-candidate & common, aligned & common, independent \\
\midrule
\multirow{3}{*}{\shortstack[l]{3 cand.\\(decoy best $11.6\%$)}}
 & \est{DM}  & $13.6\%$ & $10.0\%$ & $16.3\%$ \\
 & \est{DR}  & $13.9\%$ & $13.5\%$ & $12.9\%$ \\
 & \est{IPS} & $18.1\%$ & $20.5\%$ & $16.8\%$ \\
\midrule
\multirow{3}{*}{\shortstack[l]{7 cand.\\(decoy best $4.0\%$)}}
 & \est{DM}  & $4.2\%$ & $4.0\%$  & $8.7\%$ \\
 & \est{DR}  & $5.6\%$ & $4.9\%$  & $7.1\%$ \\
 & \est{IPS} & $6.2\%$ & $11.3\%$ & $8.2\%$ \\
\bottomrule
\end{tabular}
\end{table}

\subsection{Per-dataset $\est{DM}-\est{IPS}$ selection gaps}
\label{app:rq4-perdataset}
Paired per-cell $\est{DM}-\est{IPS}$ correct-selection gap, per dataset, under all six
designs (regenerated from \texttt{full\_run}, \texttt{selection\_run},
\texttt{commonlog\_3cand} and \texttt{commonlog\_run}). These are the numbers behind
Table~\ref{tab:rq4-logger}'s sign counts and behind the observation that \dataset{Jobs} is
negative in \emph{all six} designs.

\begin{center}\small
\begin{tabular}{lrrrrrr}
\toprule
& \multicolumn{3}{c}{3-candidate slate} & \multicolumn{3}{c}{7-policy slate} \\
\cmidrule(lr){2-4}\cmidrule(lr){5-7}
dataset & per-cand. & aligned & indep. & per-cand. & aligned & indep. \\
\midrule
\dataset{Hillstrom} & $+0.080$ & $+0.120$ & $+0.007$ & $-0.022$ & $+0.022$ & $-0.111$ \\
\dataset{IHDP}      & $+0.300$ & $+0.293$ & $+0.373$ & $+0.400$ & $+0.389$ & $+0.322$ \\
\dataset{Jobs}      & $-0.160$ & $-0.113$ & $-0.187$ & $-0.244$ & $-0.367$ & $-0.300$ \\
\dataset{Lenta}     & $+0.053$ & $+0.040$ & $-0.067$ & $-0.144$ & $-0.167$ & $-0.133$ \\
\dataset{Synthetic} & $+0.140$ & $+0.187$ & $+0.153$ & $+0.233$ & $+0.222$ & $+0.300$ \\
\midrule
mean (= Table~\ref{tab:rq4-logger}) & $+0.083$ & $+0.105$ & $+0.056$ & $+0.044$ & $+0.020$ & $+0.016$ \\
\bottomrule
\end{tabular}
\end{center}

\noindent Cells are balanced within each column ($150$ per dataset on the
three-candidate slate, $90$ on the seven-policy slate), so each column's mean over the
five datasets is exactly the pooled gap reported in Table~\ref{tab:rq4-logger}. As
throughout RQ4, the full-budget point $k{=}1$ is excluded.

\medskip\noindent
The same paired per-cell contrast for \emph{normalized regret} ($\est{DM}-\est{IPS}$;
negative favours \est{DM}), which is what Section~\ref{sec:rq4} reads as the regret gap:

\begin{center}\small
\begin{tabular}{lrrrrrr}
\toprule
& \multicolumn{3}{c}{3-candidate slate} & \multicolumn{3}{c}{7-policy slate} \\
\cmidrule(lr){2-4}\cmidrule(lr){5-7}
dataset & per-cand. & aligned & indep. & per-cand. & aligned & indep. \\
\midrule
\dataset{Hillstrom} & $-0.047$ & $-0.085$ & $-0.030$ & $+0.092$ & $+0.037$ & $+0.083$ \\
\dataset{IHDP}      & $-0.213$ & $-0.297$ & $-0.330$ & $-0.206$ & $-0.306$ & $-0.266$ \\
\dataset{Jobs}      & $+0.130$ & $+0.129$ & $+0.192$ & $+0.166$ & $+0.208$ & $+0.231$ \\
\dataset{Lenta}     & $+0.010$ & $+0.0007$ & $+0.068$ & $+0.192$ & $+0.143$ & $+0.054$ \\
\dataset{Synthetic} & $-0.009$ & $-0.012$ & $-0.010$ & $-0.018$ & $-0.020$ & $-0.023$ \\
\midrule
mean                & $-0.026$ & $-0.052$ & $-0.022$ & $+0.045$ & $+0.012$ & $+0.016$ \\
\bottomrule
\end{tabular}
\end{center}

\noindent The two exact-value datasets favour \est{DM} in every design; of the three
HT-reference datasets, \dataset{Jobs} and \dataset{Lenta} favour \est{IPS} in every design
and \dataset{Hillstrom} switches sign with the slate. That split, not the logging design,
is what drives the sign reversal of the pooled regret gap between the two slates.

\begin{table}[t]\centering
\caption{RQ4 policy selection: rate of selecting the reference-best policy (the truly best on exact-value datasets, the HT-reference-best on RCT datasets; higher better), mean normalized regret (lower better), and SharpeRatio@$k{\geq}2$ (higher better; the ratio of a policy's estimated value to the standard deviation of that estimate across seeds, restricted to slates of at least two candidates, following \citet{kiyohara2024scope}). \textbf{This table uses the per-candidate (self-aligned) logging design}, in which each candidate is scored on a log built from its own score, so it compares policy--logger pairs rather than policies. Section~\ref{sec:rq4} shows the model-based-over-IPS margin is nonetheless comparable to what a shared log gives ($1.3\times$ under the common-aligned design, $0.7\times$ under the common-independent one); see Table~\ref{tab:rq4-logger} for the common-logger results, which we regard as the primary selection evidence.}
\label{tab:rq4-selection}
\begin{tabular}{lrrr}
\toprule
estimator & correct rate & norm. regret & Sharpe@$k{\geq}2$ \\
\midrule
dm & 0.656 & 0.153 & 11.909 \\
dr & 0.633 & 0.148 & 11.879 \\
switch\_dr & 0.631 & 0.149 & 11.881 \\
snips & 0.589 & 0.160 & 11.734 \\
mIPS & 0.580 & 0.172 & 11.023 \\
ips & 0.573 & 0.179 & 11.019 \\
\bottomrule
\end{tabular}
\end{table}
\begin{table}[t]\centering
\caption{RQ4 selection quality split by reference type: rate of selecting the best candidate. On exact-value datasets (\dataset{synthetic}, \dataset{IHDP}) this is the \emph{truly} best policy; on RCT-reference datasets (\dataset{Jobs}, \dataset{Hillstrom}, \dataset{Lenta}) it is the policy ranked best by the noisy HT reference. Three-candidate slate; uniform-random baseline $=1/3\approx0.33$. \textbf{Per-candidate (self-aligned) logging design} --- see Table~\ref{tab:rq4-logger}.}
\label{tab:rq4-reftype}
\begin{tabular}{lrr}
\toprule
estimator & exact-value (truly best) & RCT-reference (reference-best) \\
\midrule
dm & 0.827 & 0.542 \\
dr & 0.777 & 0.538 \\
switch\_dr & 0.773 & 0.536 \\
snips & 0.693 & 0.520 \\
ips & 0.607 & 0.551 \\
mIPS & 0.610 & 0.560 \\
\bottomrule
\end{tabular}
\end{table}

\section{Shared vs disjoint RCT reference split}
\label{app:refdep}

On the RCT datasets the reported error is relative to an HT \emph{reference estimate};
in the main runs the logged data and that reference are built from the same randomized
evaluation split. To check that this shared-split dependence does not drive the
estimator ranking, we re-ran all three RCT datasets --- \dataset{Hillstrom},
\dataset{Lenta}, and \dataset{Jobs} --- under a disjoint
$40/30/30$ split into policy-training, logged-feedback, and reference-estimation units,
computing each estimate on the logged split and scoring it against an HT reference on
the \emph{independent} reference split. Both columns below are medians of the relative
RMSE over $10$ seeds $\times$ $3$ budgets $\times$ $3$ overlaps $\times$ $3$ policies
(\texttt{make repro-rct-disjoint}; \texttt{results/refdep\_run/}). We print four decimals
because the differences \emph{within} a column are of order $0.001$:

\begin{center}
\small
\begin{tabular}{llrr}
\toprule
dataset & estimator & shared ref. & disjoint ref. \\
\midrule
\dataset{Hillstrom} & \est{DM}      & 0.0312 & \textbf{0.0595} \\
                    & \est{DR}      & 0.0353 & 0.0635 \\
                    & \est{IPS}     & \textbf{0.0305} & 0.0631 \\
                    & \est{SNIPS}   & 0.0313 & 0.0622 \\
\midrule
\dataset{Lenta}     & \est{DM}      & 0.0421 & \textbf{0.0624} \\
                    & \est{DR}      & \textbf{0.0405} & 0.0648 \\
                    & \est{IPS}     & 0.0435 & 0.0653 \\
                    & \est{SNIPS}   & 0.0433 & 0.0681 \\
\midrule
\dataset{Jobs}      & \est{DM}      & 0.0845 & \textbf{0.1391} \\
                    & \est{DR}      & 0.0765 & 0.1515 \\
                    & \est{IPS}     & 0.0759 & 0.1628 \\
                    & \est{SNIPS}   & \textbf{0.0752} & 0.1469 \\
\bottomrule
\end{tabular}
\end{center}

\textbf{Both columns come from this reduced design, not from the main run.} The
``shared ref.'' column is the matched control \emph{within} the $40/30/30$ experiment
--- same $30\%$ logged split and same three-budget subset, with the reference computed
on the shared logged units --- so the two columns are comparable to each other but
\emph{not} to Table~\ref{tab:rq1-accuracy}, whose \dataset{Hillstrom} column uses the
main $50/50$ split and all six budgets. That is why the absolute levels here
(e.g.\ \dataset{Hillstrom} \est{DM} $0.0312$) sit above the main-run values
(\est{DM} $0.022$): the logged split is smaller and the budget grid differs. Only the
within-appendix comparison is meaningful.

\textbf{The reference, not the estimator, sets the level.} Moving to the disjoint
reference raises relative RMSE for every estimator on every dataset --- by $48\%$
(\dataset{Lenta}), $65\%$ (\dataset{Jobs}) and $91\%$ (\dataset{Hillstrom}) for
\est{DM}, and by more for \est{IPS} --- and that level shift is roughly $2$--$6\times$
the spread across estimators \emph{within} either column. The disjoint reference is a
noisier HT estimate computed on fewer units, and most of the increase is reference
variance rather than estimator error.

\textbf{Which estimator looks best depends on which reference is used.} \est{DM} is
never best under the shared reference --- a weighting or DR estimator wins on each
dataset (\est{IPS} $0.0305$ on \dataset{Hillstrom}, \est{DR} $0.0405$ on \dataset{Lenta},
\est{SNIPS} $0.0752$ on \dataset{Jobs}) --- and is best on \emph{all three} under the
disjoint one. \est{IPS} moves the opposite way: first on \dataset{Hillstrom} and second
on \dataset{Jobs} (within $0.0007$ of \est{SNIPS}) under the shared reference, third,
third and last under the disjoint one. Note that \est{IPS} is worst of the four only on
\dataset{Jobs} ($0.1628$); on \dataset{Hillstrom} the worst is \est{DR} ($0.0635$) and on
\dataset{Lenta} \est{SNIPS} ($0.0681$), so this is a shift in the whole ordering rather
than an \est{IPS}-specific penalty.

\textbf{How much this experiment can carry.} The direction matches the mechanism we
proposed: the HT reference is itself an inverse-propensity construction, so on shared
units it shares sampling noise with \est{IPS} and the two agree partly for reasons
unrelated to accuracy; removing that sharing costs weighting estimators their apparent
edge. But the differences within a column are $0.001$--$0.009$ on \dataset{Hillstrom} and
\dataset{Lenta}, this is three datasets, and the reference-variance term dominates the
level, so we report the result as \emph{consistent with} that mechanism rather than as an
isolation of it. (An earlier five-seed version of this table circulated with different
values and a sharper reading; the ten-seed run reported here supersedes it.) What the
experiment does establish is the negative conclusion we rely on: the choice of reference
moves every estimator's measured error by more than the estimators differ from each
other, so RCT-reference comparisons cannot resolve the estimator ordering. That is why we
decline to read the \est{IPS} column of Table~\ref{tab:rq1-accuracy} as a ranking, and it
is also why these RMSE values should be read as OPE error plus reference-estimation
variance, not as absolute estimator error. On the model-based side the check is
reassuring in the direction that matters for the paper's headline: the shared-split
construction did not manufacture the model-based advantage --- the cleaner disjoint
design produces a stronger one.

\end{document}